\documentclass[10pt,twocolumn,letterpaper]{article}

\usepackage[pagenumbers]{cvpr} 

\usepackage[dvipsnames]{xcolor}

\usepackage{graphicx}
\usepackage{amsmath}
\usepackage{amssymb}
\usepackage{booktabs}
\usepackage{multirow}
\usepackage{cuted}
\usepackage{capt-of}
\usepackage{colortbl}
\usepackage[ruled,vlined]{algorithm2e}
\usepackage[hypcap=false]{caption}

\definecolor{cvprblue}{rgb}{0.21,0.49,0.74}
\usepackage[pagebackref,breaklinks,colorlinks,citecolor=cvprblue]{hyperref}

\usepackage[capitalize]{cleveref}
\crefname{section}{Sec.}{Secs.}
\Crefname{section}{Section}{Sections}
\Crefname{table}{Table}{Tables}
\crefname{table}{Tab.}{Tabs.}

\def\paperID{3487} 
\def\confName{CVPR}
\def\confYear{2024}

\title{Semantic Human Mesh Reconstruction with Textures}

\author{
\parbox{\linewidth}{\centering
		 Xiaoyu Zhan$^1$, Jianxin Yang$^1$, Yuanqi Li$^1$, Jie Guo$^1$, Yanwen Guo$^{1*}$, and Wenping Wang$^{2}$\\
		$^1$ Nanjing University \quad $^2$ Texas A\&M University\\
		{\tt \small{\{zhanxy,jianxin-yang,yuanqili\}@smail.nju.edu.cn, 
      \{guojie,ywguo\}@nju.edu.cn,  wenping@tamu.edu}}
      \textcolor{magenta}{\url{https://zhanxy.xyz/projects/shert}}
    }
}

\begin{document}
\maketitle

\begin{strip}\centering
\vspace{-15mm}
\includegraphics[width=\textwidth]{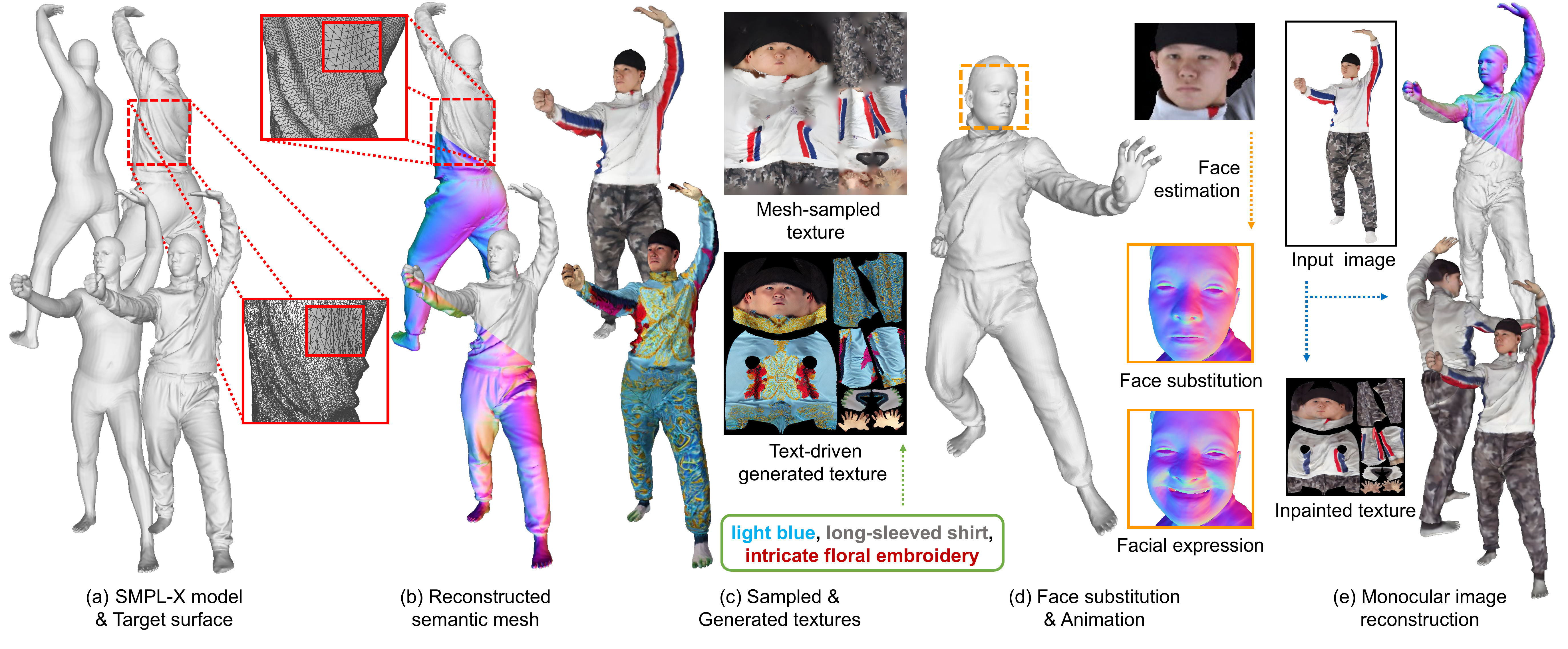}
\vspace{-9mm}
\captionof{figure}{\textbf{Semantic Human mEsh Reconstruction with Textures (SHERT):} (a) Given a target surface and the corresponding semantic guider, (b) SHERT reconstructs a detailed semantic model, which has stable UV unwrapping and skinning weights with high-quality triangle meshes. (c) It can either sample a texture map from the target surface or generate from the text prompts. (d) Based on our semantic representation, SHERT allows for high-precision facial reconstruction and animation of the body, face, and hands. (e) Moreover, SHERT is capable of inferring a fully textured avatar from a monocular image. }
\label{fig:feature-graphic}
\end{strip}

\begin{abstract}

The field of 3D detailed human mesh reconstruction has made significant progress in recent years. However, current methods still face challenges when used in industrial applications due to unstable results, low-quality meshes, and a lack of UV unwrapping and skinning weights. In this paper, we present SHERT, a novel pipeline that can reconstruct semantic human meshes with textures and high-precision details. SHERT applies semantic- and normal-based sampling between the detailed surface (\eg mesh and SDF) and the corresponding SMPL-X model to obtain a partially sampled semantic mesh and then generates the complete semantic mesh by our specifically designed self-supervised completion and refinement networks. Using the complete semantic mesh as a basis, we employ a texture diffusion model to create human textures that are driven by both images and texts. Our reconstructed meshes have stable UV unwrapping, high-quality triangle meshes, and consistent semantic information. The given SMPL-X model provides semantic information and shape priors, allowing SHERT to perform well even with incorrect and incomplete inputs. The semantic information also makes it easy to substitute and animate different body parts such as the face, body, and hands. Quantitative and qualitative experiments demonstrate that SHERT is capable of producing high-fidelity and robust semantic meshes that outperform state-of-the-art methods.

\end{abstract}    
\section{Introduction}

Recovering highly realistic details and textures of a human mesh from monocular images is crucial for various applications such as gaming, movies, cartoons, VR, virtual try-on, and digital avatars. Current approaches \cite {pifu:iccv:2019,pifuhd:cvpr:2020,icon:2022:cvpr,econ2023,pamir:2021:pami,arch:cvpr:2020,arch++:iccv:2021,2k2k:2023:cvpr, dif:2023:cvpr, tex2shape:iccv:2019, VBR:2018:cvpr, detailedhuman:2018:3dv, Monoclothcap:2020:3dv, CAR:2023:cvpr, shen2023xavatar} primarily focus on the recovery of the geometric details that are associated with images, but their results are not yet practical for real-world applications. Recent advancements in parametric and explicit clothing models \cite{xiang2021modeling, smplicit:cvpr:2021, bcnet:eccv:2020, ReEF:2022:cvpr, feng2022capturing, qiu2023rec} have shown promise in clothing reconstruction. However, they have limitations in accurately fitting different geometries and details. Implicit reconstruction approaches \cite{pifu:iccv:2019,pifuhd:cvpr:2020,icon:2022:cvpr,econ2023,pamir:2021:pami,2k2k:2023:cvpr, dif:2023:cvpr} excel at capturing clothing details but perform poorly in hands and face reconstruction. They may also generate incomplete and geometrically inseparable results.

In this work, our goal is to reconstruct fully textured semantic human meshes through given detailed surfaces and corresponding semantic guiders. Our semantic human mesh is complete, animatable, and edit-friendly for both users and designers. It ensures that each vertex has deterministic semantic information and predefined skinning weights. This allows for easy substitution and animation of different body parts such as the face, body, and hands. The semantic information also guarantees stable and reasonable UV unwrapping, which is advantageous for editing and image-based texture generation. 

Specifically, we propose SHERT, which generates a semantic human mesh from the detailed surface and its corresponding SMPL-X \cite{SMPL-X:2019} model and optionally infers textures from images or colored surfaces. SHERT has four main processes. \textbf{1) Sampling}, SHERT applies semantic and normal-based sampling to obtain a partially sampled semantic mesh based on the input detailed surface and SMPL-X model. We subdivide the original SMPL-X model to better capture the human details. \textbf{2) Completion}, a self-supervised network is proposed to complete the partially sampled mesh. The network works in the 2D UV domain, which converts the 3D completion task into a 2D inpainting task. \textbf{3) Refinement}, the origin image and front-back normal maps are used to enhance the geometry details. \textbf{4) Texture}, our semantic human mesh is projected to the origin image in order to generate the partial texture map. Then we adapt the diffusion framework for text-driven partial texture inpainting and generation. The powerful generation ability of pre-trained diffusion model enables textures with rich clothing details and clear facial expressions.

Extensive experiments have been conducted on both datasets and in-the-wild images. The quantitative and qualitative results show the robustness and superior performance of SHERT in reconstructing high-fidelity semantic human meshes and generating various high-resolution textures. 

In summary, the main contributions of this work include four-fold.
\begin{itemize}
  \item We introduce SHERT, a novel pipeline to reconstruct high-quality semantic human meshes from the detailed 3D surfaces, represented either explicitly as meshes, or implicitly as signed distance field (SDF). SHERT is also capable of predicting robust and fully textured avatars with high-fidelity faces from monocular images.
  \item We propose a semantic- and normal-based sampling method (SNS) and a self-supervised mesh completion network to achieve non-rigid 3D surface registration. The approach has the capability to process incomplete and inaccurate inputs by leveraging SMPL-X human priors.
  \item We present a self-supervised mesh refinement network working in the UV domian. It utilizes the images and front-back normal maps to improve the geometric mesh details.
  \item We use a diffusion model to infer high-resolution human textures from input images. The model can also accomplish text-driven texture inpainting and generation. 
\end{itemize}

\section{Related Work}

\begin{figure*}[t]
  \centering
   \includegraphics[width=1\linewidth]{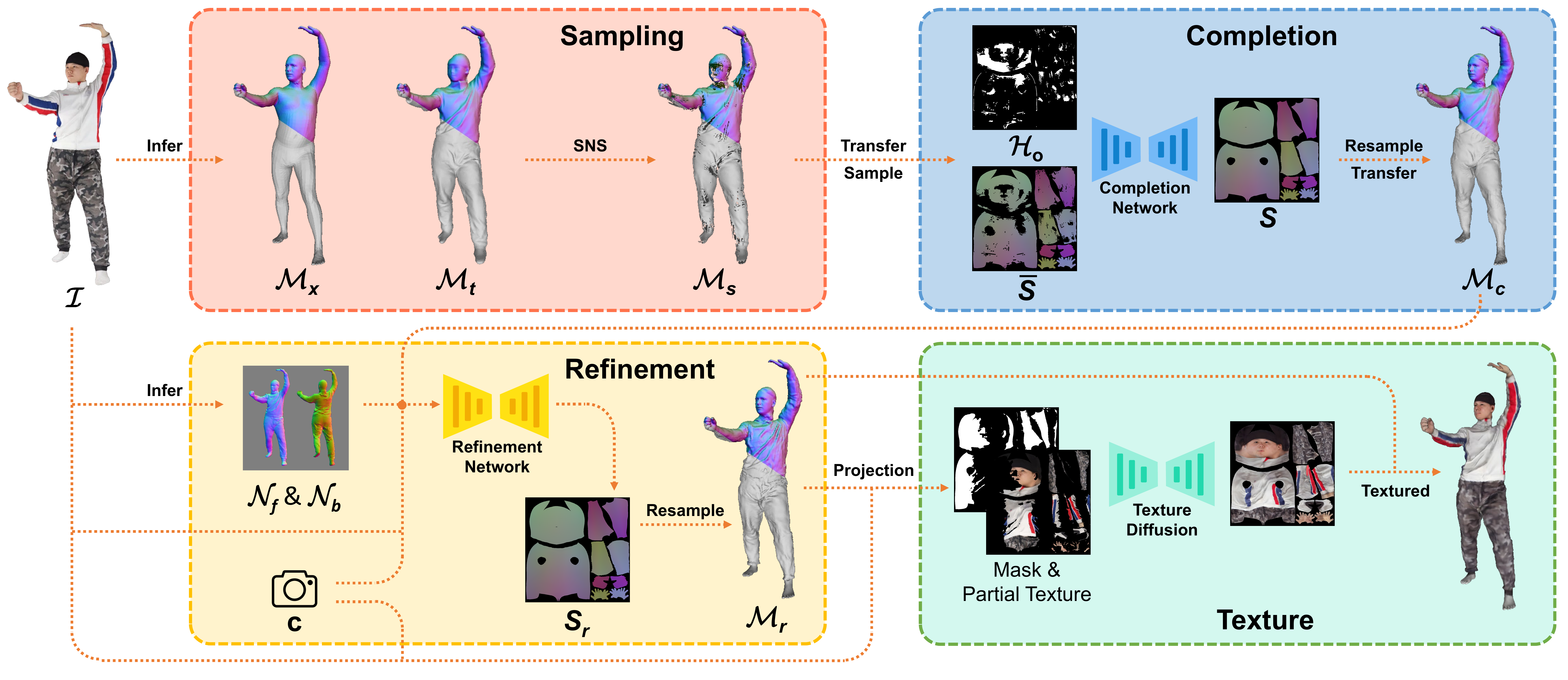}
   \vspace{-8mm}
   \caption{\textbf{Overview of SHERT for monocular image reconstruction}. Given an RGB image $\mathcal{I}$, SHERT first infers the detailed mesh $\mathcal{M}_t$ and corresponding sub-SMPLX model $\mathcal{M}_x$. It then applies SNS to obtain the partial semantic mesh $\mathcal{M}_s$ (in Sec. \ref{sec:m2}). The Completion Net infers $\mathcal{M}_c$ by filling the UV holes in $\mathcal{M}_s$ (in Sec. \ref{sec:m3}). Image and normal maps are utilized to generate $\mathcal{M}_r$, which contains sharper geometry details (in Sec. \ref{sec:m4}). Finally, SHERT uses a diffusion model to achieve text-driven texture inpainting and generation (in Sec. \ref{sec:m5}). } 
   \vspace{-5mm}
   \label{fig:pipeline}
\end{figure*}

\subsection{Monocular 3D Human Reconstruction}
\textbf{Clothed Human Reconstruction.} Most of the current CNN-based monocular 3D clothed human shape estimation methods can be divided into two categories: explicit-based \cite{VBR:2018:cvpr,alldieck2019learning,hmd:2019:cvpr,detailedhuman:2018:3dv,tex2shape:iccv:2019,bcnet:eccv:2020,hmd_pami:2021:tpami} and implicit-based \cite{pifu:iccv:2019, pifuhd:cvpr:2020, smplicit:cvpr:2021, pamir:2021:pami, icon:2022:cvpr, econ2023,2k2k:2023:cvpr,arch:cvpr:2020,arch++:iccv:2021,dif:2023:cvpr,phorhum:2022:cvpr,CAR:2023:cvpr,tech:2024:3dv, shen2023xavatar} approaches, based on their representations. Explicit-based approaches usually infer the 3D offsets on top of the parametric human model \cite{SMPL:2015, SMPL-X:2019, scape:pr:2017, ghum:cvpr:2020}. However, these methods are difficult to apply to flexible human topologies and cannot capture details well. Implicit-based approaches have the advantage of representing arbitrary 3D clothed human shapes that are free from the limitations of parametric human models. There are also some works \cite{pamir:2021:pami,icon:2022:cvpr,econ2023,CAR:2023:cvpr,corona2023structured,tech:2024:3dv} that mix the implicit and explicit representations to achieve the detailed and robust 3D clothed human reconstruction. But both implicit-based and mixed methods still cannot maintain the stability of the human body shape well, often resulting in problems such as blurring and missing local body parts in the predicted results. Some methods \cite{bcnet:eccv:2020, smplicit:cvpr:2021,feng2022capturing,xiang2021modeling,ReEF:2022:cvpr,qiu2023rec} propose additional parametric or implicit clothing models to fit loose clothes. These clothing models often struggle to accurately capture the details present in the images. Nevertheless, they are still useful for generating rough approximations of the clothing. Despite considerable progress made in monocular human reconstruction capture, there are still certain constraints, especially in terms of reliability, availability, and user-friendliness.

\noindent
\textbf{Texture Prediction.} 
Previous works \cite{mir2020learning,pifu:iccv:2019, arch:cvpr:2020, arch++:iccv:2021, phorhum:2022:cvpr, dinar:2023:iccv,corona2023structured,tech:2024:3dv, shen2023xavatar} have also focused on predicting image-based human textures. However, the current emphasis of implicit-based methods \cite{pifu:iccv:2019, arch:cvpr:2020, arch++:iccv:2021,phorhum:2022:cvpr,corona2023structured,tech:2024:3dv} is still on predicting vertex colors, which cannot be easily converted into usable texture maps for industrial applications. DINAR \cite{dinar:2023:iccv} proposes neural textures for modeling human avatars and achieves texture inpainting by utilizing the diffusion framework. Nevertheless, the neural texture is highly coupled with human avatars, and the resolution is relatively low.

\noindent
\textbf{Face Reconstruction.} In recent years, parameterized face models \cite{3DMM:1999:top,BFM:2009,facewarhouse:2014:tvcg,FLAME:2017:tog} have been widely used in monocular face reconstruction algorithms \cite{3DDFA:2016:cvpr, deng2019accurate, mgcnet:2020:eccv,DECA:Siggraph2021,EMOCA:CVPR:2021}, contributing to the success of achieving realistic high-quality reconstruction results. In light of the incorporation of FLAME \cite{FLAME:2017:tog} into SMPL-X \cite{SMPL-X:2019}, we can utilize high-quality facial reconstruction results from previous works to enhance the accuracy and realism of the SMPL-X based human body reconstruction results.

\subsection{Non-rigid 3D Registration}
Non-rigid 3D surface registration approaches \cite{Nicp:2007,snr:2015:cgf,fastRNRR:cvpr:2020,RPTS,clusterReg:2022:tvcg,SVR,mda:2023:sig,Large:2021:sig} usually compute a deformation that aligns a source surface with a target surface \cite{RegSurvey:2022:eg}. SHERT aims to transfer the semantic information of parametric human models to the corresponding detailed surfaces through registration. However, existing methods mainly focus on the precise registration of the source and target, while the target detailed human surfaces in real-world tasks often have incomplete and erroneous data.

\section{Method}
SHERT is capable of reconstructing a high-fidelity fully textured semantic human mesh based on a pair of detailed 3D surface and corresponding SMPL-X \cite{SMPL-X:2019} model. Our results have high robustness and will not result in missing body parts in despite of the incomplete inputs. Furthermore, the semantic mesh ensures that each vertex has deterministic semantic information and predefined skinning weights, making it possible to replace and animate the human face, body, and hands. SHERT can also generate realistic human textures from texts and images. These features make it easy for users and designers to use or further edit our results. To achieve this, SHERT first subdivides the SMPL-X model to improve the accuracy in capturing human details (in Sec. \ref{sec:m1}) and then applies the semantic- and normal-based sampling to obtain a partially sampled mesh (in Sec. \ref{sec:m2}). We complete the partial result using a self-supervised mesh completion network (in Sec. \ref{sec:m3}), and finally enhance the geometry details through refinement (in Sec. \ref{sec:m4}). In addition, SHERT uses a diffusion model for text-driven human texture inpainting and generation (in Sec. \ref{sec:m5}).

\subsection{Subdivided SMPL-X model}
\label{sec:m1}
In this work, we use SMPL-X as the semantic guider. SMPL-X is a parameterized body model that combines SMPL \cite{SMPL:2015} with the FLAME \cite{FLAME:2017} head model and the MANO \cite{MANO:2017} hand model. It is parameterized with shape and pose, has $10,475$ vertices and $54$ joints, including joints for the neck, jaw, eyeballs, and fingers. The SMPL-X model provides basic skinning weights and corresponding semantic information for each vertex. 

SHERT defines the sub-SMPLX based on SMPL-X since $10,475$ vertices are not sufficient to accurately represent the details of the human body and clothing, such as facial details and clothing wrinkles.  Taking into consideration both the expressiveness of the model and computational cost , we apply the mid-point subdivision algorithm twice on a standard SMPL-X model (with eyeballs removed) to obtain the sub-SMPLX with $149,921 $ vertices and $299,712$ faces. 

\begin{figure}[t]
  \centering
   \includegraphics[width=1\linewidth]{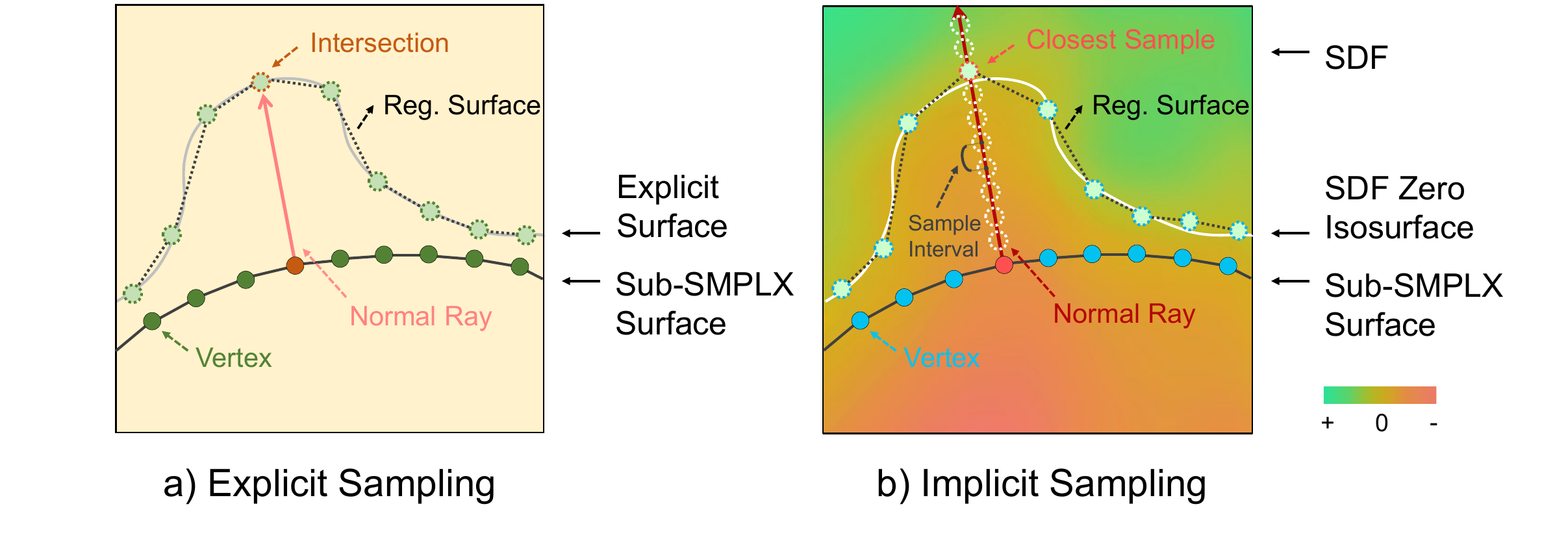}
   \vspace{-7mm}
   \caption{\textbf{Explicit and implicit sampling}. a) SNS shoots a ray from the starting vertex along the vertex normal to locate the intersection point with the target surface. b) SNS takes samples at fixed intervals along the vertex normal ray and search for the point that is closest to the zero isosurface.}
   \vspace{-6mm}
   \label{fig:registration_example}
\end{figure}

\subsection{Semantic- and Normal-based Sampling (SNS)}
\label{sec:m2}
Our objective is to accurately predict the non-rigid deformation between a source surface and a target surface. Given a source surface $\mathcal{M}_x$, which is represented as a sub-SMPLX in our work, and the target surface $M_t$, we learn a mapping function $D: \mathcal{M}_x^{149921 \times 3} \rightarrow \mathcal{M}_s^{149921 \times 3}$ such that $\mathcal{M}_s$ can be aligned with $\mathcal{M}_t$ through the learned mapping in geometry, and ignoring the incorrect parts.

\begin{figure}[t]
  \centering
   \includegraphics[width=1\linewidth]{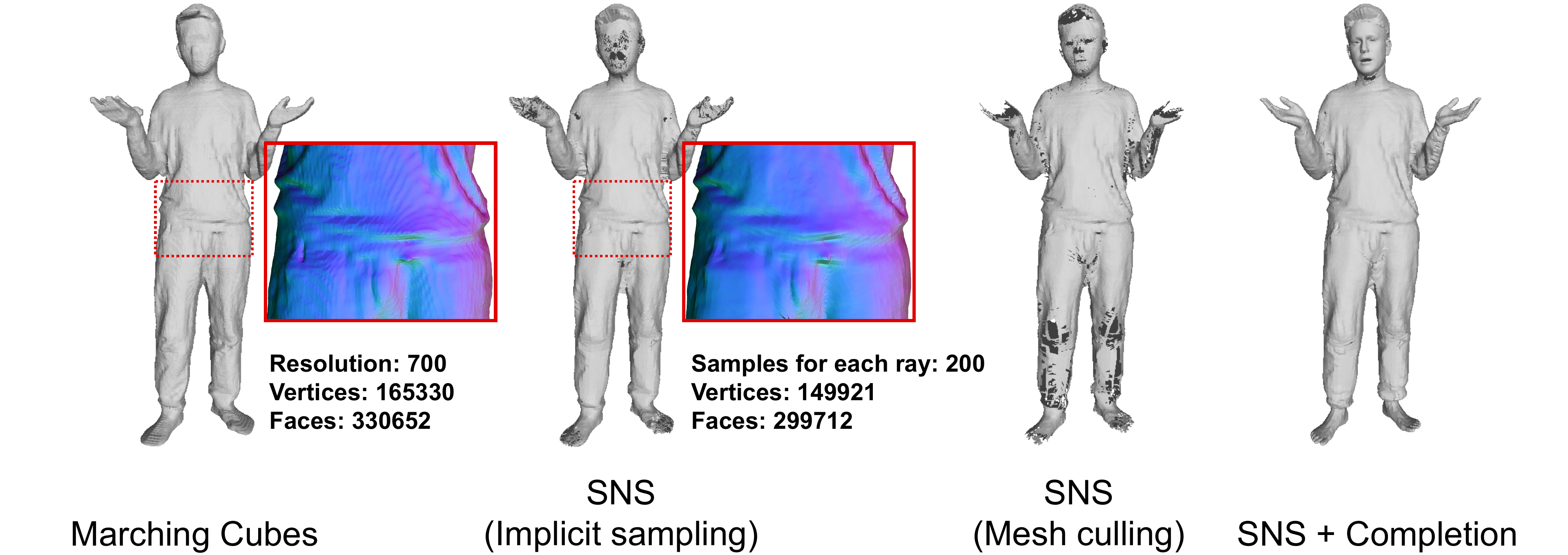}
   \vspace{-5mm}
   \caption{\textbf{SNS in the implicit field}.  We compare the performance of SNS and the Marching Cubes algorithm \cite{lorensen1998marching} in the implicit field predicted by the multi-view PIFu \cite{pifu:iccv:2019}. The results show that our SNS and completion network correctly reconstruct the implicit field and obtain a smoother surface compared to Marching Cubes.}
   \vspace{-5mm}
   \label{fig:sns}
\end{figure}

Specifically, we obtain a partially sampled mesh through a sampling scheme based on the vertex normals $N^{149921\times3}$. Starting from a point on $\mathcal{M}_x$, we search for the intersection point with the target surface $\mathcal{M}_t$ along the vertex normal ray. We can extend the sampling scheme to implicit surfaces with constant step Ray Marching \cite{perlin1989hypertexture, jensen2001raymarching}, shown in Fig. \ref{fig:registration_example} and Fig. \ref{fig:sns}. 

It should be noted that SNS does not always return satisfactory results since the ray may not intersect with the target surface. We need to locate and label the vertices that have failed to register. In addition, due to the large geometric differences and the possibility of incorrect alignment, the sampling scheme may retrieve incorrect results, which should also be detected. In order to remove incorrect points and triangle meshes from the sampled results, we calculate $\theta$ (the angle between the normal vectors of the sampled triangle mesh and the corresponding sub-SMPLX triangle mesh), $s$ (the area ratio between the sampled triangle mesh and the corresponding sub-SMPLX triangle mesh), and $r$ (the edge ratio between the longest and shortest edges of the sampled triangle mesh) for each triangle mesh in the sampled result. We then perform mesh culling by removing low-quality triangle meshes that have at least one indicator exceeding the threshold among the three mentioned above. Finally, we perform a connectivity check on the processed sampled mesh and remove the sets with a number of connected triangle meshes less than $g$. By default, we set $\theta=2$, $s=3$, $r=3$, and $g=500$.

We found that culling meshes only in the current pose would result in unreasonable deformations in human animation tasks. Therefore, SNS repeats the processing above in the canonical space. This ensures that we remove almost all unreasonable sampling results and obtain the partial semantic mesh $\mathcal{M}_s$.

\subsection{Self-supervised Mesh Completion}
\label{sec:m3}
In Sec. \ref{sec:m2}, the vertices that failed to register have resulted in holes in $M_s$. Currently, there are many mesh completion algorithms \cite{liepa2003filling, kazhdan2006poisson, kraevoy2005template, kazhdan2013screened, ifnet:2020:cvpr} that can fill in these mesh holes. However, these algorithms cannot be well adapted to our semantic reconstruction task due to the strong constraints on the number and relative positions of vertices in our representation in Sec. \ref{sec:m1}. 

\begin{figure}[t]
  \centering
   \includegraphics[width=1\linewidth]{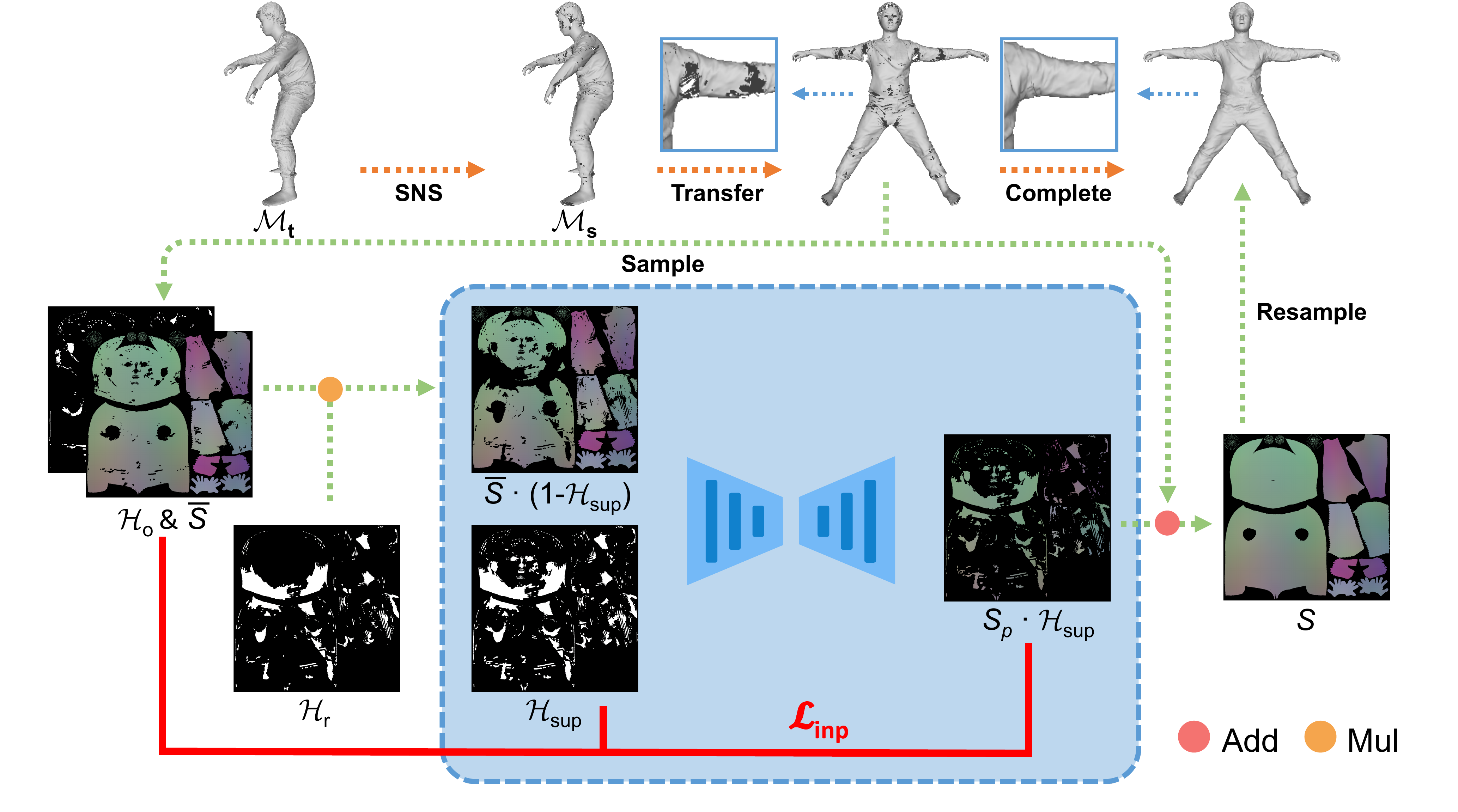}
   \vspace{-7mm}
   \caption{\textbf{Completion network}. The completion network transforms the result of SNS into canonical space and predicts the holes in the UV domain.}
   \vspace{-5mm}
   \label{fig:completion_network}
\end{figure}

\begin{figure}[t]
  \centering
   \includegraphics[width=1\linewidth]{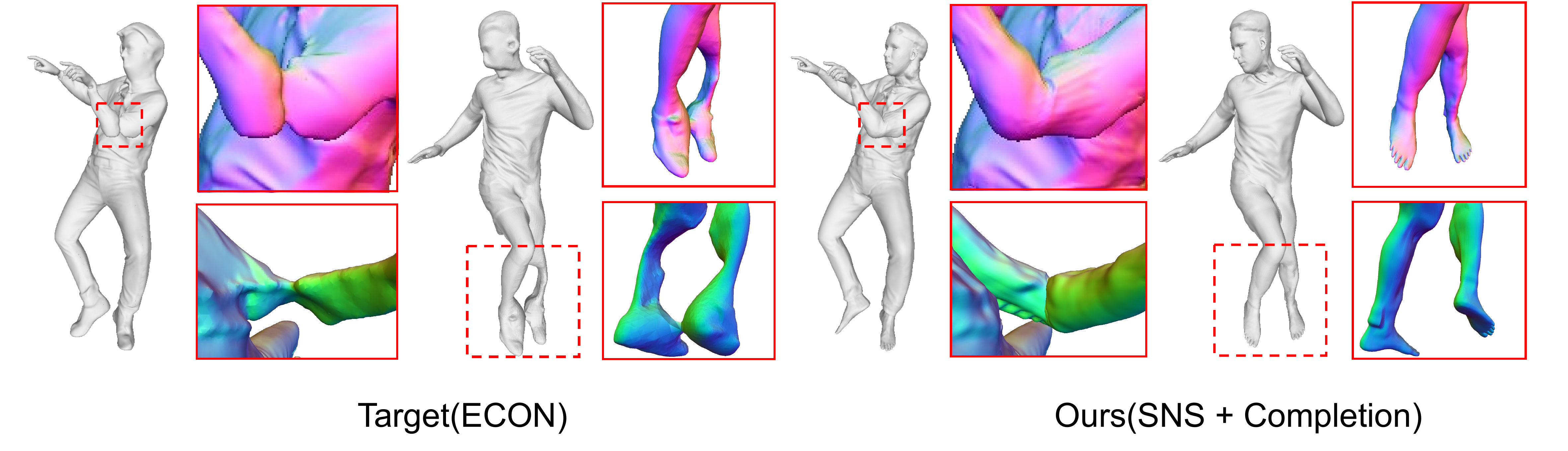}
   \vspace{-7mm}
   \caption{\textbf{The robustness of SNS and Completion Network}. We present our complete meshes, which are reconstructed using the predictions of ECON \cite{econ2023} from in-the-wild images. Distinguishing itself from previous registration methods, SHERT has the capability to effectively process scenarios where the inputs are incomplete, inaccurate, or contain errors by leveraging SMPL-X human priors.}
   \vspace{-3mm}
   \label{fig:completion_robustness}
\end{figure}

Therefore, we transform the partially sampled mesh into the UV domain \cite{deng2018uv,tex2shape:iccv:2019, ooh:2020:cvpr} according to the semantic information. As a result, we convert the mesh completion task in 3D space into an inpainting task on a 2D image. Then, we design a self-supervised completion network that can fill in the holes of the partially sampled mesh. As shown in Fig. \ref{fig:completion_network}, by adding a random hole mask $\mathcal{H}_r$ on the partial UV position map, we can obtain trainable pairs, which has a consistent distribution as the missing parts of the sampled mesh. The manually masked parts can provide the network with consistent supervision information as the input parts. This completion network is capable of generating robust meshes while maintaining both mesh quality and semantic consistency in the final results (refer to Fig. \ref{fig:completion_robustness}).

To further decrease the learning difficulties, we transform all incompletely sampled meshes to canonical space when facing the diverse range of poses and clothing styles. We believe that this approach has the potential to mitigate the challenges in problem-solving, as completion is no longer affected by pose and body shape. Meanwhile, since the sampled points are located on the normal ray of the sub-SMPLX vertices, we represent the deformation relative to sub-SMPLX as an offset based on the vertex normal. The transformed UV position map $\overline{S}$ can be presented as:

\begin{equation} \label{equ:1}
\overline{d} = \frac{S_{sample} - S_{pose}}{N_{pose}},
\end{equation}

\begin{equation} 
\overline{S} = S_{cano} + N_{cano} \cdot \overline{d},
\end{equation}

\noindent
where $S_{sample}$ denotes the partial UV position map of the sampled mesh. $S_{pose}$ and $S_{cano}$ refer to the UV position maps of the original posed sub-SMPLX and the canonical sub-SMPLX respectively. $N_{pose}$ and $N_{cano}$ are the corresponding UV normal maps. 

The complete UV position map $S$ and the input combined hole mask $\mathcal{H}_{sup}$ are calculated by:

\begin{equation}
\mathcal{H}_{sup} = \mathcal{H}_r \cdot (1 - \mathcal{H}_o) + \mathcal{H}_o,
\end{equation}

\begin{equation} 
S = S_{cano} + N_{cano} \cdot [~d \cdot \mathcal{H}_o + \overline{d} \cdot (1 - \mathcal{H}_o)],
\end{equation}

\noindent 
where $\mathcal{H}_r$ and $\mathcal{H}_o$ denote the randomly added hole mask  and the hole mask generated by SNS respectively. $d \in \mathbb{R}^{H\times W\times 3}$ is the estimated displacement UV map. The complete mesh $\mathcal{M}_c$ is resampled from $S$ and then transformed to the original pose space.

The primary loss used in our completion network is as follows:

\begin{equation} 
S_p = S_{cano} + N_{cano} \cdot [d \cdot \mathcal{H}_{sup} + \overline{d} \cdot (1 - \mathcal{H}_{sup})],
\end{equation}

\begin{equation}
\mathcal{L}_{inp} = \frac{\left\|(S_p - \overline{S}) \cdot (\mathcal{H}_{sup} - \mathcal{H}_o)\right\|_2^2}{\sum_{i,j}(\mathcal{H}_{sup} - \mathcal{H}_o)_{i,j}}.
\end{equation}

We find that completing the face, hands, and feet is much more difficult than other parts of the body, so we use the corresponding parts of sub-SMPLX for replacement. Furthermore, SHERT allows for the use of FLAME-based methods such as EMOCA \cite{EMOCA:CVPR:2021} to replace the facial region in our completion results, enhancing the accuracy and realism of the facial details.

\subsection{Self-supervised Mesh Refinement}
\label{sec:m4}
After completing the mesh, we have already reconstructed a semantic mesh with extremely high levels of detail. However, some geometric details may be lost during the SNS and completion process. Therefore, we design an additional self-supervised refinement network to further optimize the mesh's details (\eg cloth wrinkles, subtle deformations of body movements) using the image and normal maps. The input normal maps can be obtained either by rendering the scanned model or by using prediction networks. Our refinement network follows a U-Net \cite{ronneberger2015unet} architecture and is capable of predicting a displacement UV map, denoted as $z \in \mathbb{R}^{H\times W\times 3}$, in order to optimize the complete mesh $\mathcal{M}_c$. Since the input image and normal maps are not in the UV domain, we utilize a separate network to extract the image domain features and then project them to UV domain.

\begin{figure}
  \centering
   \includegraphics[width=1\linewidth]{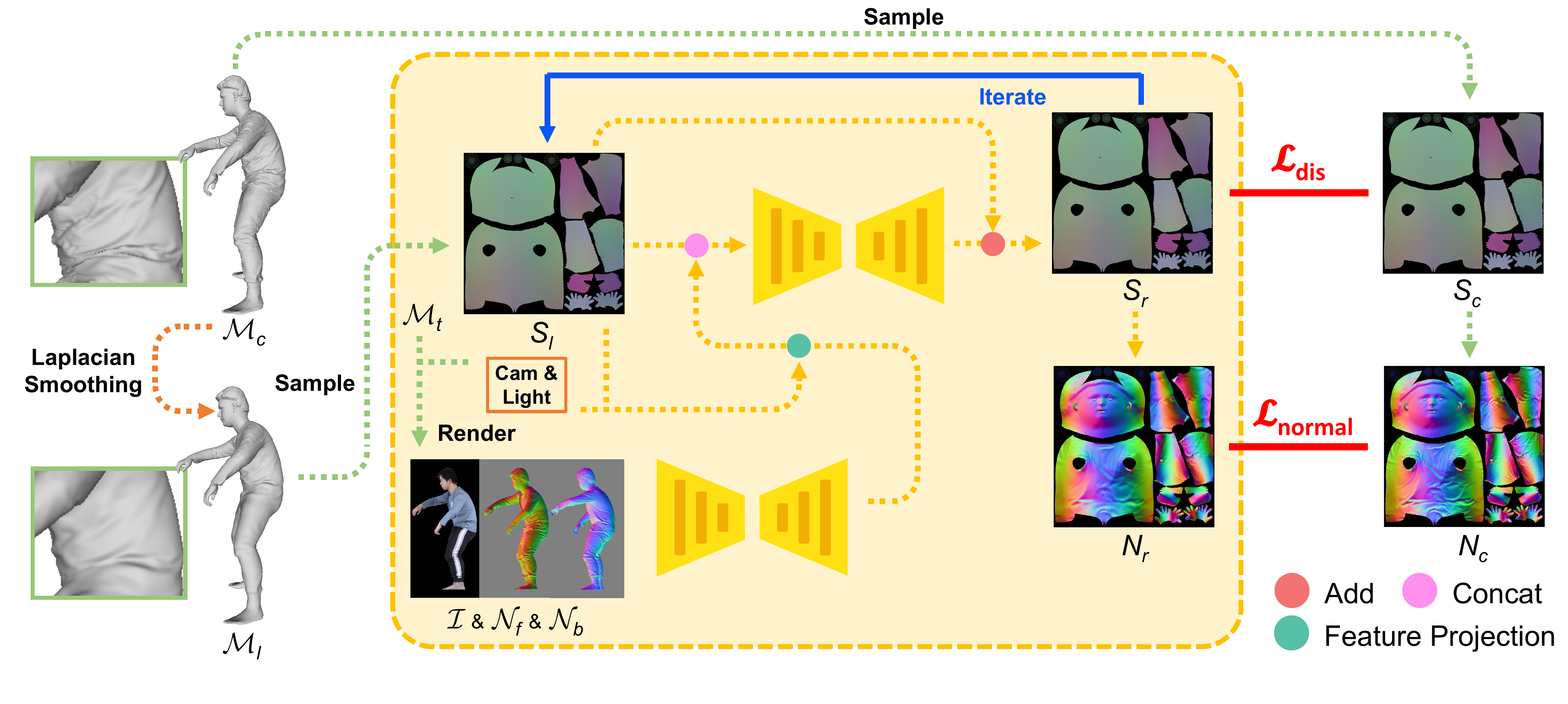}
   \vspace{-8mm}
   \caption{\textbf{Refinement network}. The features extracted from the image and front-back normal maps are projected to the UV domain. These projected features are subsequently combined with the input UV position map to generate a refined mesh.}
   \vspace{-5mm}
   \label{fig:refine_network}
\end{figure}

\begin{figure*}[t]
  \centering
   \includegraphics[width=1\linewidth]{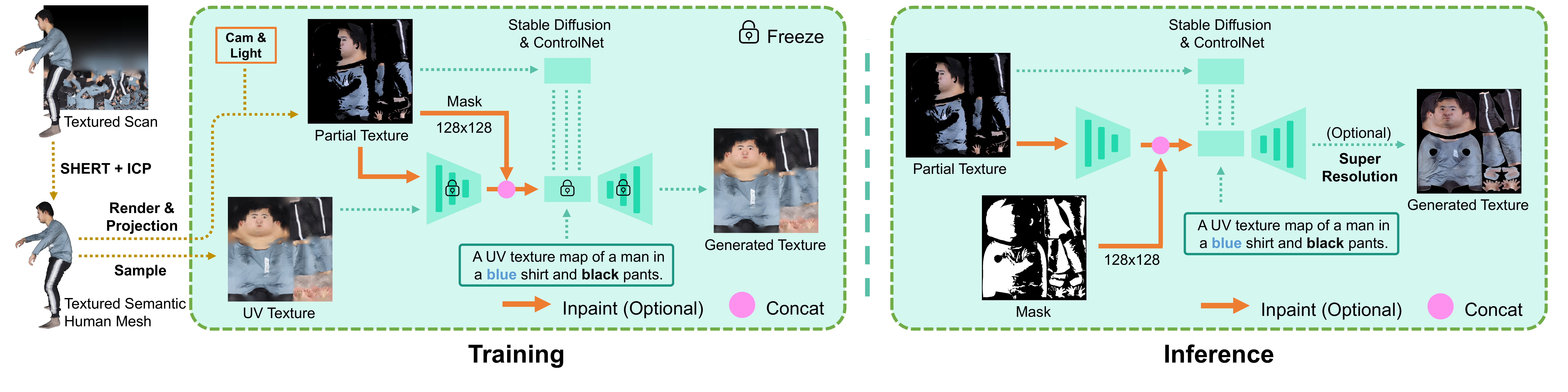}
   \vspace{-10mm}
   \caption{\textbf{Texture diffusion}. We use two strategies to finetune the diffusion model for text-driven texture repainting and inpainting separately. During inference, instead of just using random noise, the encoded partial texture map and mask can be manually added to the latent features to preserve the original information of the input. Additionally, we use the super resolution method Real-esrgan \cite{Real-esrgan:2021:iccv}, which can specifically optimize both the body and face, to further enhance the final outputs.}
   \vspace{-6mm}
   \label{fig:diffusion_network}
\end{figure*}

In order to guide the network in learning the potential correspondence between the input features and mesh details, we apply Laplacian Smoothing \cite{field1988laplacian} on the existing complete meshes, as shown in Fig. \ref{fig:refine_network}. The smoothed mesh $\mathcal{M}_l$ is used as input to the network, and the original mesh $\mathcal{M}_c$ is used for supervision. 

Given the image $\mathcal I$, front-view normal map $\mathcal{N}_f$, back-view normal map $\mathcal{N}_b$, and the corresponding complete UV position map $S_l$, the projected feature $F$ is typically represented as

\begin{equation} 
F = \mathcal{P}(\mathcal{F}(\mathcal{I}, \mathcal{N}_f, \mathcal{N}_b), S_l, c),
\end{equation}

\noindent 
where $c$ represents the camera parameters, $\mathcal{F}$ denotes the image domain feature extraction network, and $\mathcal{P}$ is the projection function that can project the image domain features to the UV domain according to the camera parameters and the 3D coordinates of points on $S_l$.

The UV position map $S_r$ of the refined mesh $\mathcal{M}_r$ is given by

\begin{equation} 
S_r = S_l + N_{l} \cdot z,
\end{equation}

\noindent
where $z \in \mathbb{R}^{H\times W\times 3}$ represents the predicted displacement UV map of the refine network, $S_l$ and $N_l$ are the UV position map and normal UV map of $\mathcal{M}_l$ respectively.

The loss we used in refine network can be represented as

\begin{equation} \label{equ:9}
\mathcal{L}_{dis} = MSE(z - \frac{S_c - S_l}{N_l}),
\end{equation}

\begin{equation} 
\mathcal{L}_{normal} = MSE(N_r - N_c),
\end{equation}

\noindent
where $S_c$ is the UV position map of $\mathcal{M}_c$. $N_r$ and $N_c$ denote the normal UV maps of $\mathcal{M}_r$ and $\mathcal{M}_c$ respectively.

\begin{figure}[b]
  \centering
   \vspace{-5mm}
   \includegraphics[width=1\linewidth]{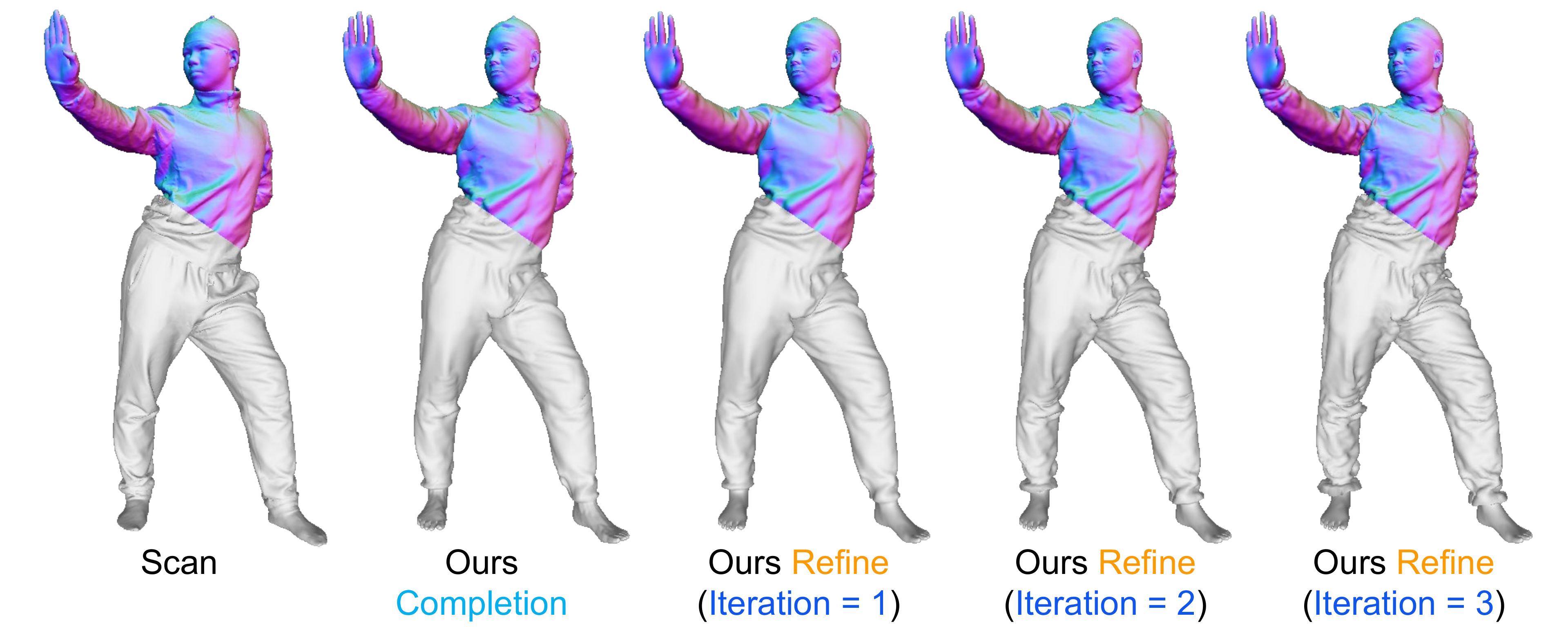}
   \vspace{-7mm}
   \caption{\textbf{The ablation results (with face substitution)}. We present the results after completion and refinement. With an increasing number of refinement iterations, the details of the mesh will be enhanced. Please zoom-in to see more details.}
   \label{fig:refine_iteration}
\end{figure}

Moreover, the refine network can be iteratively employed by users to enhance the mesh details according to their desired level (refer to Fig. \ref{fig:refine_iteration}).

\subsection{Text-driven Texture Inpainting and Generation}
\label{sec:m5}
In Sec. \ref{sec:m4}, all the reconstructed meshes share the same semantic information from SMPL-X. This means that when the models are unfolded to the UV domain, vertices with the same semantic information will be projected to a fixed UV position, resulting in a stable texture map that allows for repainting and transferring textures. To perform texture inpainting on incomplete human body textures and generate high-quality human body textures driven by text, we fine-tune the Stable Diffusion model \cite{Stablediffusion:2022} with ControlNet \cite{controlnet:2023:iccv}, similar to the approach taken by Dinar \cite{dinar:2023:iccv}.

As our reconstructed semantic mesh closely matches the scanned model's geometry, we can use the ICP algorithm \cite{icp} to register the vertex colors from the scanned model onto our result. This enables us to convert the texture maps into the SMPL-X format. As shown in Fig. \ref{fig:diffusion_network}, we obtain a partial texture map from the input image based on the semantic mesh and camera parameters and hope to generate the invisible parts. ControlNet enable us to add conditional inputs (partial texture maps) to the generation of the diffusion network at a low cost. We use two strategies to finetune the diffusion model, distinguished by whether to take the encoded partial texture map and mask as the additional inputs \cite{stable-diffusion-inpainting}. During the inference process, we can directly input partial texture maps into the ControlNet and generate various results by adding random noise. Alternatively, we can enhance the preservation of existing information in images by concatenating the encoded partial texture maps and masks with the latent features. Although Stable Diffusion has not worked with UV-parameterized images before, the well-designed UV parameterization keeps the shapes of the face, body, and limbs stable, ensuring a learnable space for the model. The texture map for the body and limbs focuses more on color and patterns rather than shapes, also resulting in excellent outcomes when cropped with a mask.

\section{Experiments}

\subsection{Datasets and Networks}

\begin{table}[b]
  \vspace{-4mm}
  \centering
  \scalebox{0.61}{
  \begin{tabular}{@{}c|c|c|c|c|c|c@{}}
    \toprule[2pt]
    \multirow{2}{*}{Method} & \multicolumn{3}{c|}{CAPE\cite{CAPE:CVPR:20}} & \multicolumn{3}{c}{THuman2.0\cite{thuman2:2021:cvpr}} \\
    \cline{2-7}
    & P2S$\downarrow$ & Chamfer$\downarrow$ & Normal$\downarrow$ & P2S$\downarrow$ & Chamfer$\downarrow$ & Normal$\downarrow$ \\
    \hline
    \hline
    PIFu\cite{pifu:iccv:2019} $^\ast$ & 2.1137 & 1.6537 & 0.0755 & 2.5493 & 2.3640 & 0.1042 \\
    PIFuHD\cite{pifuhd:cvpr:2020} & 3.7846 & 3.5787 & 0.1002 & 3.0772 & 3.1808 & 0.1207 \\
    PaMIR\cite{pamir:2021:pami} $^\ast$ & 1.4520 & 1.2241 & 0.0610 & 1.5439 & 1.3311 & 0.1102 \\
    ICON\cite{icon:2022:cvpr} & 0.8855 & 0.8609 & \textbf{0.0347} & \textbf{1.0361} & 1.0874 & 0.0607 \\
    ECON\cite{econ2023} & 0.9403 & 0.9386 & 0.0374 & 1.1304 & 1.2081 & 0.0661 \\
    2K2K\cite{2k2k:2023:cvpr} $^\dag$ & - & - & - & 2.5342 & 2.6165 & 0.1030 \\
    \hline
    \hline
    \textbf{Ours (ICON-comp)} & \textbf{0.8550} $\color{blue}\uparrow$ & \textbf{0.8107} $\color{blue}\uparrow$ & 0.0359 & 1.0459 & 1.0465 $\color{blue}\uparrow$ & 0.0604  $\color{blue}\uparrow$\\
    \textbf{Ours (ICON-refine)} & 0.8633  & 0.8112  & 0.0380  & 1.0442 $\color{red}\uparrow$ & 1.0468  & \textbf{0.0603} $\color{red}\uparrow$ \\
    \cline{1-7}
    \textbf{Ours (ECON-comp)} & 0.8561 $\color{blue}\uparrow$ & 0.8242 $\color{blue}\uparrow$ & 0.0378 & 1.1255 $\color{blue}\uparrow$ & 1.1420 $\color{blue}\uparrow$ & 0.0672 \\
    \textbf{Ours (ECON-refine)} & 0.8581  & 0.8144 $\color{red}\uparrow$ & 0.0398  & 1.0630 $\color{red}\uparrow$ & \textbf{1.0430} $\color{red}\uparrow$ & 0.0649 $\color{red}\uparrow$ \\
    \bottomrule[2pt]
    \end{tabular}}
    \vspace{-2mm}
    \caption{\textbf{Quantitative evaluation for monocular image reconstruction}. We evaluate the performance of our completion results (comp) and refinement results (refine) by comparing them with state-of-the-art methods. $\ast$ methods are re-implemented in \cite{icon:2022:cvpr} to ensure a fair comparison. $\dag$ method has only been tested with human-facing-forward images. $\color{blue}\uparrow$ and $\color{red}\uparrow$ indicate the improvement achieved through completion and refinement, respectively.} 
  \label{tab:compare1}
\end{table}

\begin{figure*}[ht]
  \centering
   \includegraphics[width=1\linewidth]{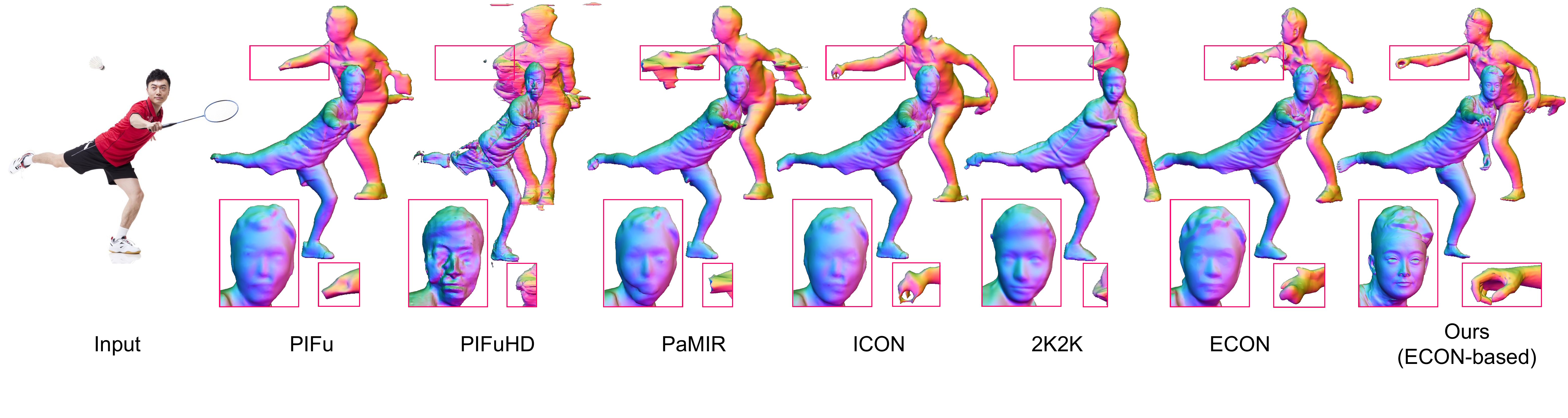}
   \vspace{-13mm}
   \caption{\textbf{Qualitative comparison for monocular image reconstruction on in-the-wild image}. For each method, we present two views of the reconstructed results. SHERT demonstrates the ability to handle challenging poses while providing clear details of facial and hand geometry.}
   \vspace{-4mm}
   \label{fig:compare_geo}
\end{figure*}

\begin{figure*}[ht]
  \centering
   \includegraphics[width=1\linewidth]{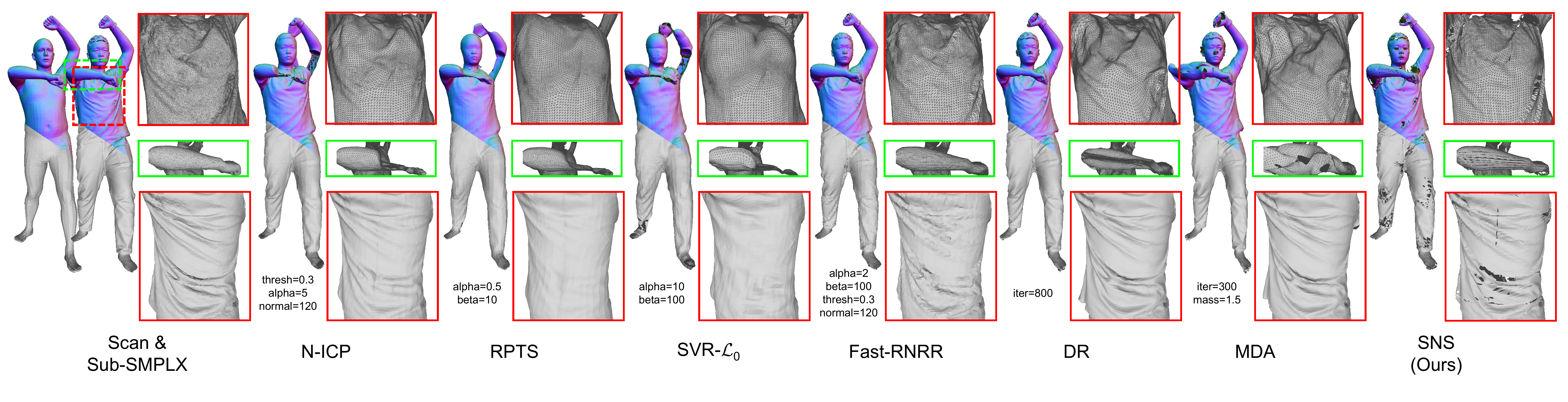}
   \vspace{-10mm}
   \caption{\textbf{Qualitative comparison for registration on THuman2.0}. We compare the registration quality of various methods including N-ICP \cite{Nicp:2007}, RPTS \cite{RPTS}, SVR-$\mathcal{L}_0$ \cite{SVR}, Fast-RNRR \cite{fastRNRR:cvpr:2020}, DR \cite{Large:2021:sig}, MDA \cite{mda:2023:sig} and ours SNS. The holes present in our result are the eliminated faces, as described in Sec. \ref{sec:m2}. The results indicate that SNS exhibits excellent performance in terms of model details, mesh quality, and registration robustness. The quantitative comparisons are shown in Tab. \ref{tab:regis_compare}.}
   \vspace{-4mm}
   \label{fig:registration}
\end{figure*}

\noindent
\textbf{Training data.}
The completion and refinement networks are trained using the first 499 scans of THuman2.0 \cite{thuman2:2021:cvpr}. During the training of the completion network, we randomly choose one of the remaining 498 masks as a random hole mask. To enrich the inputs for the refinement network, we rotate the meshes every 60 degrees, resulting in a total of 2994 different orientations. We utilized the ICP \cite{icp} algorithm to transfer the color of the THuman2.0 scans to the vertices of our completed result, thereby generating 499 UV textures for training and obtaining 2994 visible UV masks from the rotated meshes. The complete UV textures and visible UV masks are randomly combined as inputs for our texture diffusion.

\noindent
\textbf{Testing data.} We conduct quantitative and qualitative evaluations on CAPE \cite{CAPE:CVPR:20}, THuman2.0, and in-the-wild images. We use CAPE-NFP \cite{icon:2022:cvpr} (100 samples with 3 viewpoints for each), and the last 27 subjects of THuman2.0 scans (6 viewpoints, each differing by 60 degrees). 

\begin{table}
  \centering
  \scalebox{0.7}{
  \begin{tabular}{@{}c|c|c|c|c|c|c@{}}
    \toprule[1.5pt]
    Method & GPU & P2S$\downarrow$ & Chamfer$\downarrow$ & G-avg$\uparrow$ & $\theta$\textless$30^{\circ}\downarrow$ & Time \\
    \hline
    \hline
    N-ICP \cite{Nicp:2007} &  - & 0.213 & 0.163 & 0.506 & 55.2 & 7m 23s \\
    RPTS \cite{RPTS} &  - & 0.488 & 0.360 & 0.565 & 48.3 & 1m 55s \\
    SVR-$\mathcal{L}_0$ \cite{SVR} &  - & 0.404 & 0.296 & 0.531 & 53.5 & 1h 23m 32s \\
    Fast-RNRR \cite{fastRNRR:cvpr:2020} & - & 0.115 & 0.097 & 0.597 & 31.1 & 1m 4s \\
    DR \cite{Large:2021:sig} & $\surd$ & 0.339 & 0.347 & 0.581 & 47.7 & 16m 17s \\
    MDA \cite{mda:2023:sig} & $\surd$ & 0.671 & 0.731 & 0.587 & 48.5 & 4m 48s \\
    \hline
    \hline
    \textbf{Ours (SNS)} & - & \textbf{0.107} & \textbf{0.078} & \textbf{0.729} & \textbf{17.3} & \textbf{23s} \\
    \textbf{Ours (Comp)} & $\surd$ & 0.139 & 0.167 & 0.662 & 28.9 & 27s (23 + 4) \\
    \bottomrule[1.5pt]
  \end{tabular}}
  \vspace{-1mm}
  \caption{\textbf{Quantitative evaluation for registration on THuman2.0}. We test all the methods on the first subject of THuman2.0. Following the previous researches \cite{frey1999surface,li2022deep}, we adapt G-avg as a method for evaluating the mesh quality. We also report the metric $\theta$\textless$30^{\circ}$, which denotes the percentage of triangle meshes in the given mesh that have an angle less than 30 degrees. Additionally, we present the metrics for our complete mesh.} 
  \vspace{-7mm}
  \label{tab:regis_compare}
\end{table}

\noindent
\textbf{Networks.} The completion net and refinement net are both trained for 100 epochs with a learning rate of $1\times10^{-6}$. The resolutions of the input and output data, including the UV position maps, images, masks and front-back normal maps, are all $1024\times1024\times3$. During inference, the ECON's predicted front-back normals ($512\times512\times3$) are upsampled to $1024\times1024\times3$ using bilinear interpolation. The texture diffusion is trained for 1 epoch with a learning rate of $2\times10^{-5}$. The sampler of texture diffusion is DDIM \cite{ddim:2020:iclr}. We use 30 steps by default and infer the texture UV map with a resolution of $1024\times1024\times3$. All the networks are trained on three NVIDIA RTX 3090 GPUs.

\begin{figure}
  \centering
   \includegraphics[width=1\linewidth]{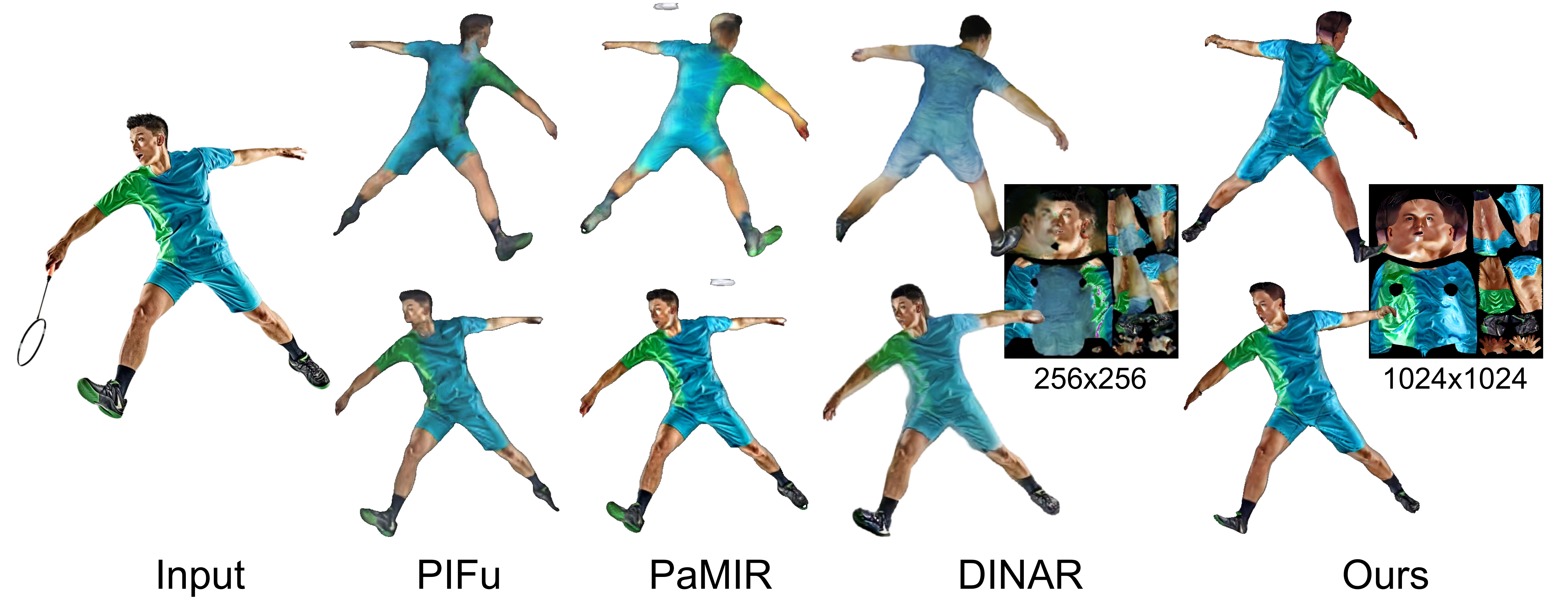}
   \vspace{-8mm}
   \caption{\textbf{Qualitative comparison for texture prediction on in-the-wild image}. We display the front and back view rendering results for each method. Since PIFu \cite{pifu:iccv:2019} and PaMIR \cite{pamir:2021:pami} predict vertex colors, we only exhibit the texture maps of DINAR \cite{dinar:2023:iccv} and SHERT (ours). Please zoom-in to see more details.}
   \vspace{-4mm}
   \label{fig:texture}
\end{figure}

\subsection{Evaluation}

\noindent
\textbf{Quantitative comparisons.} We conduct quantitative comparisons with mainstream state-of-the-art monocular image reconstruction approaches in Tab. \ref{tab:compare1}. As in previous work \cite{pifuhd:cvpr:2020,icon:2022:cvpr,econ2023,2k2k:2023:cvpr}, we report the point-to-surface Euclidean distance (P2S, cm), the Chamfer Distance (cm), and the Normals difference (L2). To ensure a fair comparison, PIFu$^\ast$ \cite{pifu:iccv:2019} and PaMIR$^\ast$ \cite{pamir:2021:pami} are re-implemented and re-trained on THuman2.0, using the same settings as ICON \cite{icon:2022:cvpr}. The ground-truth SMPL/SMPL-X models are provided for evaluation. However, PIFu \cite{pifu:iccv:2019}, PIFuHD \cite{pifuhd:cvpr:2020} and 2K2K \cite{2k2k:2023:cvpr} do not utilize the parametric body priors, which may result in subpar performance. In Tab. \ref{tab:compare1}, ``ICON-comp" refers to the completion result achieved by leveraging ICON's \cite{icon:2022:cvpr} prediction, while ``ECON-refine" denotes the refinement mesh obtained using ECON's \cite{econ2023} result and the predicted front-back normal maps. Additionally, we evaluate the registration quality of SNS against state-of-the-art non-rigid registration methods, as presented in Tab. \ref{tab:regis_compare}

\noindent
\textbf{Qualitative comparisons.}
We demonstrate a comparison between SHERT and state-of-the-art methods using in-the-wild images, with a focus on monocular image reconstruction (refer to Fig. \ref{fig:compare_geo}) and texture prediction (refer to Fig. \ref{fig:texture}). Additionally, we compare the registration quality of our SNS with state-of-the-art non-rigid registration approaches on Thuman2.0 (refer to Fig. \ref{fig:registration}).

\subsection{Limitations}
Due to the geometric limitations of SMPL-X, SHERT performs weaker in reconstructing loose clothing, shoes and hair compared to implicit-based reconstruction methods. It is also difficult to ensure consistent results of texture diffusion at the seams of UVs. See more in SupMat.

\section{Applications}

\begin{figure}[ht]
  \centering
   \includegraphics[width=1\linewidth]{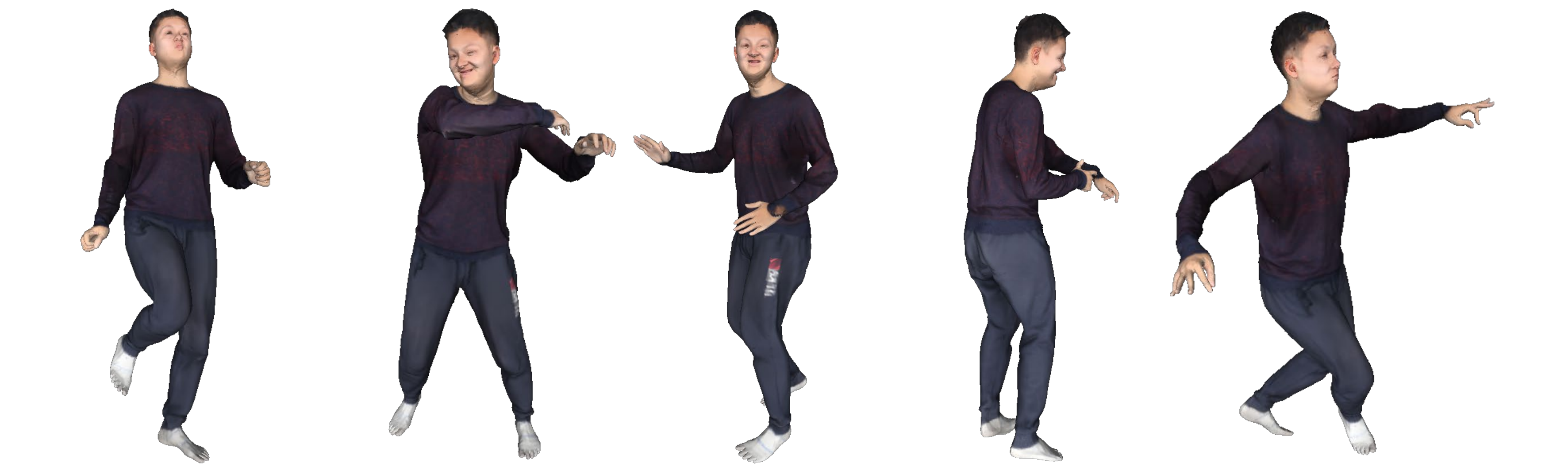}
   \vspace{-7mm}
   \caption{\textbf{Animation results}. Please zoom-in to see more details.}
   \vspace{-4mm}
   \label{fig:animation}
\end{figure}

\begin{figure}
  \centering
   \includegraphics[width=1\linewidth]{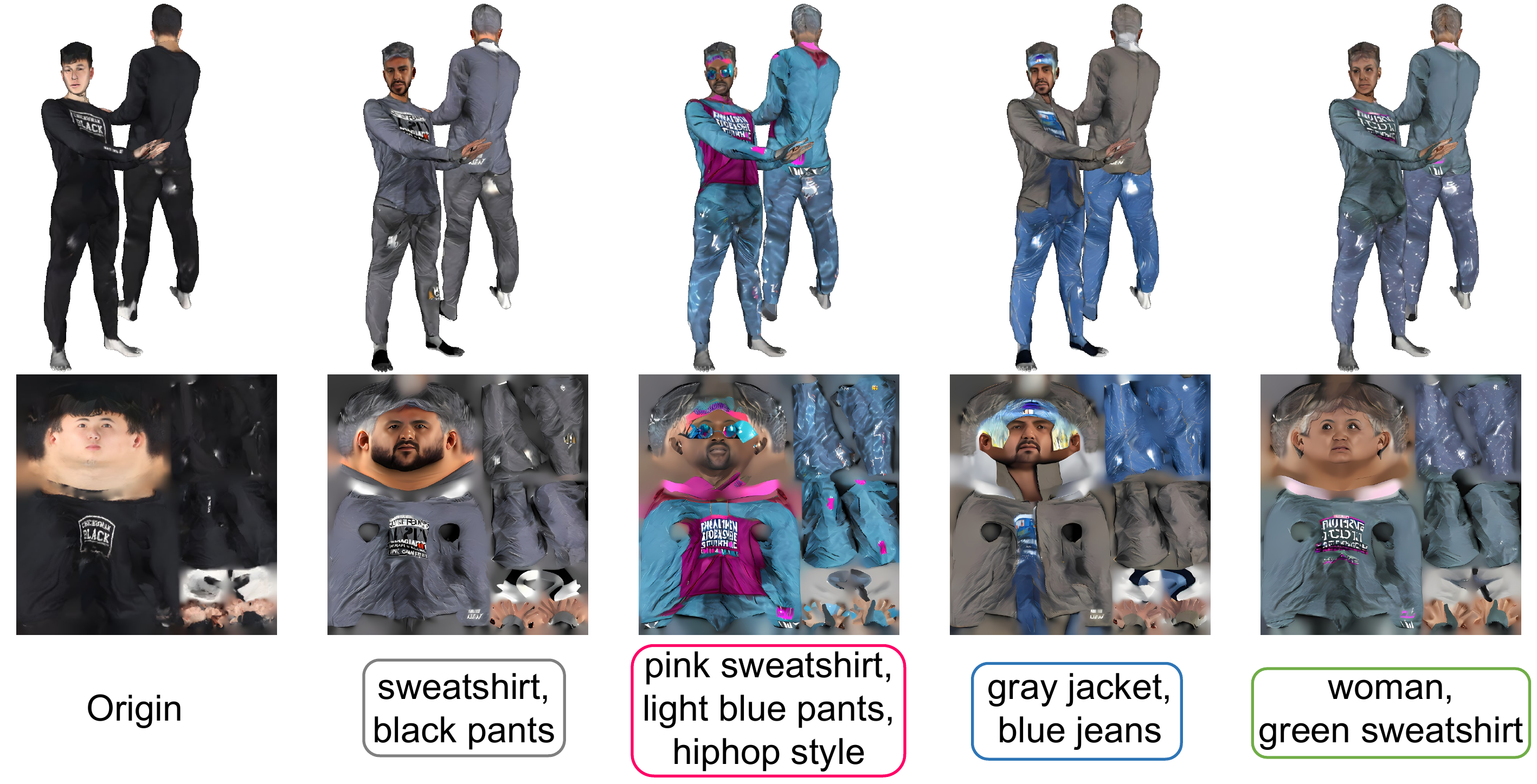}
   \vspace{-7.5mm}
   \caption{\textbf{Global texture repainting}. SHERT can repaint the texture through text prompts. Please zoom-in to see more details.}
   \vspace{-4mm}
   \label{fig:global_change}
\end{figure}

\begin{figure}
  \centering
   \includegraphics[width=1\linewidth]{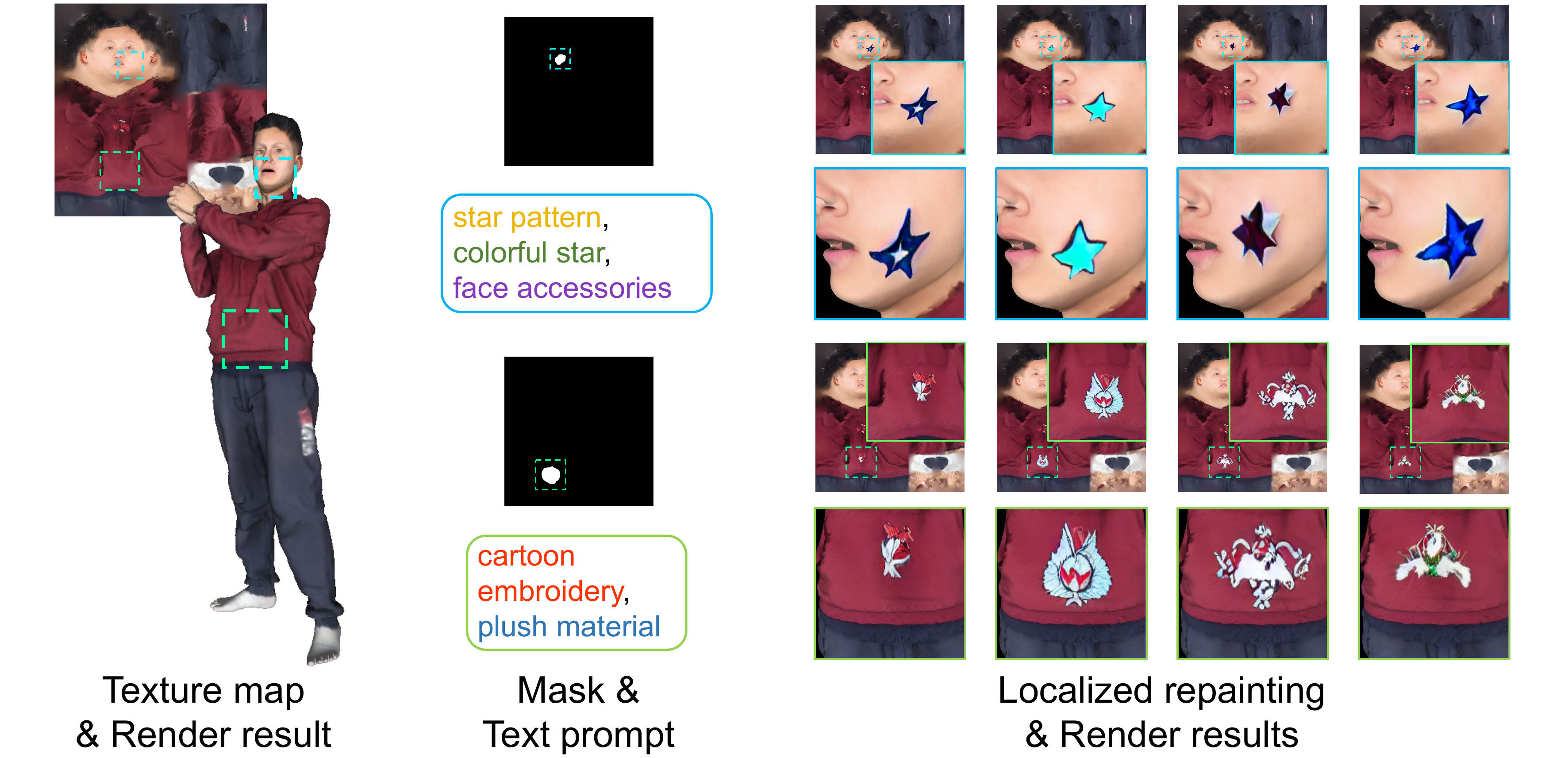}
   \vspace{-7.5mm}
   \caption{\textbf{Localized texture repainting}. SHERT can repaint the masked area through text prompts. Please zoom-in to see more details.}
   \vspace{-6.7mm}
   \label{fig:tex_change}
\end{figure}

SHERT uses the skinning weights from SMPL-X to enable animated poses, expressions, and gestures on the reconstructed mesh through LBS \cite{lbs:2000:tog} (refer to Fig. \ref{fig:animation}). It allows for both global texture repainting (refer to Fig. \ref{fig:global_change}) and the option for users to provide custom masks and text prompts for localized texture repainting (refer to Fig. \ref{fig:tex_change}).

\section{Conclusion}

We propose SHERT, which reconstructs a fully textured semantic human avatar from a detailed surface or a monocular image. It takes advantage of the geometric details of the target surface, along with semantic information and prior knowledge of the semantic guider. The reconstructed results have high-fidelity clothing details, high-quality triangle meshes, clear facial features, and complete hands geometry. SHERT is also capable of generating high-resolution texture maps with stable UV unwrapping. This approach bridges existing monocular reconstruction work and downstream industrial applications, and we believe it can promote the development of human avatars.

\section*{Acknowledgments}
This work was supported by the National Natural Science Foundation of China (62032011) and the Natural Science Foundation of Jiangsu Province (No. BK20211147).
\newpage
{
    \small
    \bibliographystyle{ieeenat_fullname}
    \bibliography{main}
}

\clearpage
\setcounter{page}{1}
\maketitlesupplementary

\setcounter{figure}{0}
\setcounter{table}{0}

\renewcommand\thesection{\Alph{section}}
\renewcommand\thetable{S\arabic{table}}
\renewcommand\thefigure{S\arabic{figure}}

\noindent
We provide additional information and experiments about our method. Below is a summary of the sections in the supplementary:
\vspace{2mm}
\begin{itemize}
\item Sec. \ref{sec:implementation} reports the implementation details of SHERT.
\vspace{1mm}
\item Sec. \ref{sec:more} displays more experiments and results.
\vspace{1mm}
\item Sec. \ref{sec:limi} discusses the limitations.
\vspace{1mm}
\item Sec. \ref{sec:appl} shows more available applications.
\end{itemize}

\section{Implementation details}
\label{sec:implementation}


\subsection{SNS}
\textbf{Algorithm}. In Alg. \ref{sns_code}, we describe the details of semantic- and normal-based sampling.

\begin{algorithm}
    \caption{Semantic- and Normal-based Sampling}
    \label{sns_code}
    \KwIn{Target surface $\mathcal{M}_t$, sub-SMPLX $\mathcal{M}_x$, query-range $r$, angle-threshold $t_{angle}$, area-threshold $t_{area}$, edge-threshold $t_{edge}$, connectivity-threshold $t_{connect}$.}
    \KwOut{Partially sampled mesh $\mathcal{M}_s$, hole mask $\mathcal{H}_o$.}
    \BlankLine
    \SetKwFunction{MeshCulling}{MeshCulling}
    \SetKwProg{Fn}{Function}{:}{}
    \Fn{\MeshCulling{$\mathcal{M}_x$, $\mathcal{M}_s$, $t_{angle}$, $t_{area}$, $t_{edge}$}}{
        List $invalid$\;
        \ForEach{face $f$ \textbf{in} $\mathcal{M}_s$}{
            Get the corresponding face $f_x$ of $\mathcal{M}_x$\;
            Compute the normal angle $angle_{pose}$ between $f$ and $f_x$\;
            Compute the area ratio $area_{pose}$ between $f$ and $f_x$\;
            Compute the edge ratio $edge_{pose}$ between the longest and the shortest edges of $f$\;
            \If{$angle_{pose}$ \textgreater $t_{angle} \lor area_{pose}$ \textgreater $t_{area} \lor edge_{pose}$ \textgreater $t_{edge}$}{
                \ForEach{vertex $p$ \textbf{in} $f$}{
                    $invalid$.append(index($p$))\;
                }
            }
        }
        $\mathcal{M}_s \gets$ DeleteFaces($invalid$)\;
        \Return{$\mathcal{M}_s$}\;
    }
    \BlankLine
    \BlankLine
    
    Copy $\mathcal{M}_x$ to $\mathcal{M}_s$\;
    List $invalid$\;
    \BlankLine
    
    \ForEach{vertex $p$ \textbf{in} $\mathcal{M}_s$}{
        $direction \gets CalNormal(p)$\;
        \If{$p$ is not inside the target surface $\mathcal{M}_t$}{
            $direction \gets direction \cdot (-1)$\;
        }
        Find the intersection point $i_p$ with $\mathcal{M}_t$ starting from $p$\ along $direction$\ within range $r$\;
        \If{$i_p$ does not exist}{
            $invalid$.append(index($p$))\;
        }
        \Else{
            $p \gets i_p$\;
        }
    }
    $\mathcal{M}_s \gets$ DeleteFaces($invalid$)\;
    \BlankLine
    $\mathcal{M}_s \gets$ MeshCulling($\mathcal{M}_x$, $\mathcal{M}_s$, $t_{angle}$, $t_{area}$, $t_{edge}$)\;
    \BlankLine
    $\mathcal{M}_x, \mathcal{M}_s \gets$ LBSToStar($\mathcal{M}_x, \mathcal{M}_s$)\;
    \BlankLine
    $\mathcal{M}_s \gets$ MeshCulling($\mathcal{M}_x$, $\mathcal{M}_s$, $t_{angle}$, $t_{area}$, $t_{edge}$)\;
    \BlankLine
    \BlankLine
    $\mathcal{M}_s \gets$ FaceConnectivityCheckAndDelete($\mathcal{M}_s$, $t_{connect}$)\;
    $\mathcal{H}_o \gets$ GenerateHoleMask($\mathcal{M}_s$)\;
    $\mathcal{M}_s \gets$ LBSBack($\mathcal{M}_s$)\;
    \BlankLine
    
    \Return{$\mathcal{M}_s, \mathcal{H}_o$}\;

\end{algorithm}

\noindent
\textbf{Skinning weights}. The skinning weight of a new vertex is the average of the skinning weights of the two neighboring vertices on the same edge.

\subsection{Network Architecture}
\begin{figure*}
  \centering
   \includegraphics[width=1\linewidth]{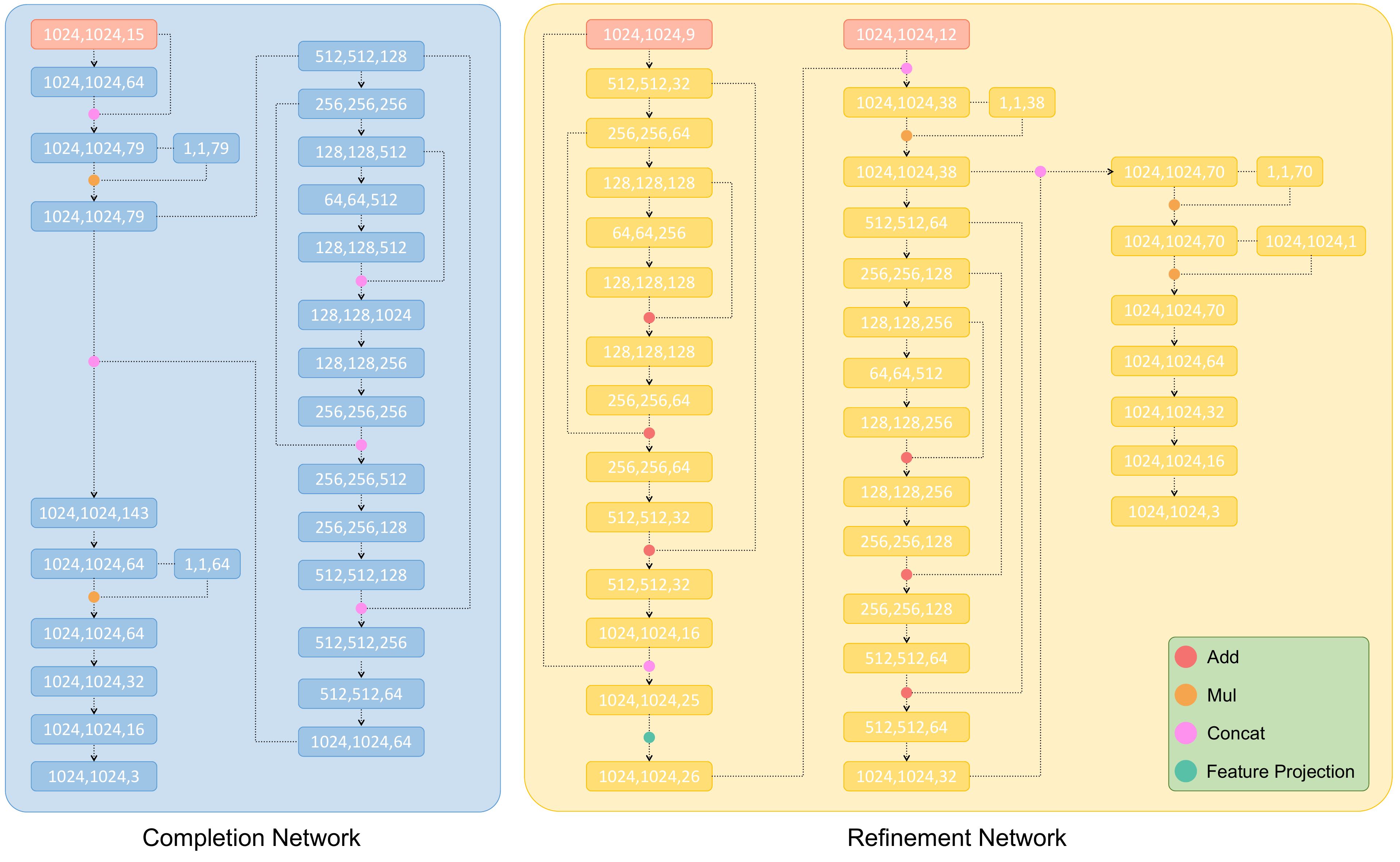}
   \vspace{-6mm}
   \caption{\textbf{Network Architecture}. The inputs are: (1) \textbf{(1024,1024,15)}, hole mask, holed uv, holed displacement uv, sub-smplx template normal uv, sub-smplx template uv. The sub-smplx template is in star pose. (2) \textbf{(1024,1024,9)}, image, front normal map, back normal map. (3) \textbf{(1024,1024,12)}, smoothed uv, normal uv, posed sub-smplx uv, posed sub-smplx normal uv. }
   \vspace{-5mm}
   \label{fig:network_arch}
\end{figure*}

The architectures of our completion network and refinement network are shown in Fig. \ref{fig:network_arch}.

\subsection{Mask Dilation in Completion}
\label{a22}
The experiments show that the vertices located at the edges of the removed triangle meshes may also be inaccurate, leading to a negative effect on the quality of the completed mesh. To tackle this problem, we dilate the hole mask, enabling the network to also fill in the surrounding areas of the eliminated triangle meshes. This leads to smoother and more realistic results.

\subsection{Refinement}
\label{a32}

\begin{figure}
  \centering
   \includegraphics[width=1\linewidth]{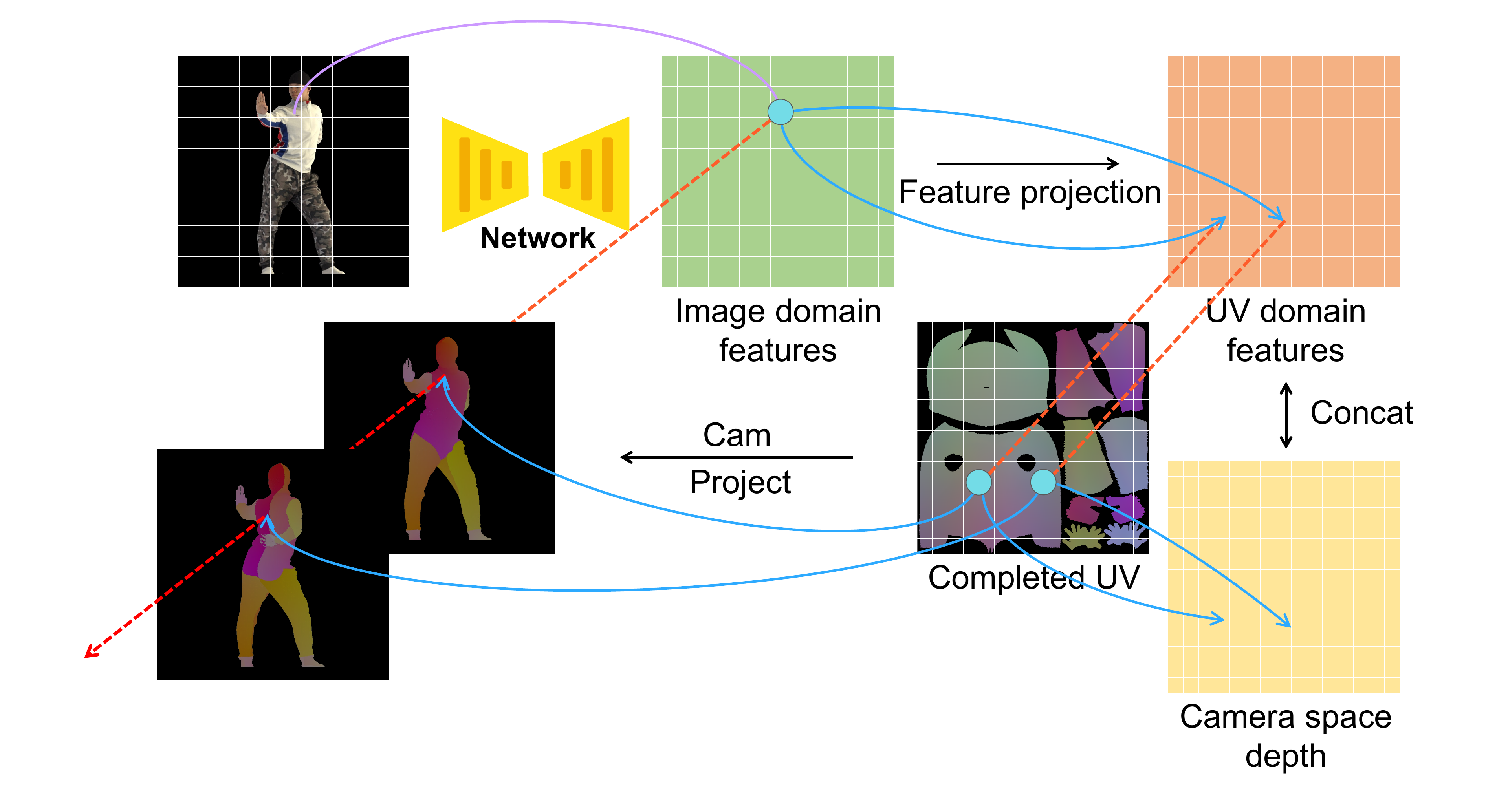}
   \vspace{-8mm}
   \caption{\textbf{Feature Projection}. Noted that the image domain features are projected to both the visible part and invisible part.}
   \label{fig:feature_projection}
\end{figure}

\noindent
\textbf{Feature Projection}. We use camera parameters to query the pixels corresponding to the current UV coordinates in the image domain to find the corresponding image domain features (refer to Fig. \ref{fig:feature_projection}). We will concatenate the depth of the projected vertex in camera space with the projected features.

\noindent
\textbf{Smoothing}. We observe that when the input mesh does not align properly with the image and normal features, the predictions made by the network tend to introduce some noise, resulting in minor protrusions on the refined mesh. In order to reduce the occurrence of such distortions, we perform Laplacian Smoothing \cite{field1988laplacian} on the input mesh before using it as input during inference. For the same reason, we also smooth the supervisory data in training.

\noindent
\textbf{Iteration}. During the iteration process of the refinement network, the input image and the front-back normal maps remain unchanged, and we only use the previous result generated by network as the new input.

\noindent
\textbf{Displacement Projection}. Actually we use the projection of displacement. Therefore, Eq. \ref{equ:1} and Eq. \ref{equ:9} in our code should be as follows:

\setcounter{equation}{0}
\begin{equation} 
\overline{d} = \frac{(S_{sample} - S_{pose}) \cdot N_{pose}}{\left\|N_{pose}\right\|_2},
\end{equation}

\setcounter{equation}{8}
\begin{equation} 
\mathcal{L}_{dis} = MSE(z - \frac{(S_c - S_l) \cdot N_l}{\left\|N_l\right\|_2}).
\end{equation}

\noindent
\textbf{Rendered images}. We render images of THuman2.0 using orthogonal cameras. Generally, this does not consistent with the real-world images and may have an impact on performance in wild tests.

\subsection{Face Substitution}
\begin{figure}
  \centering
   \includegraphics[width=1\linewidth]{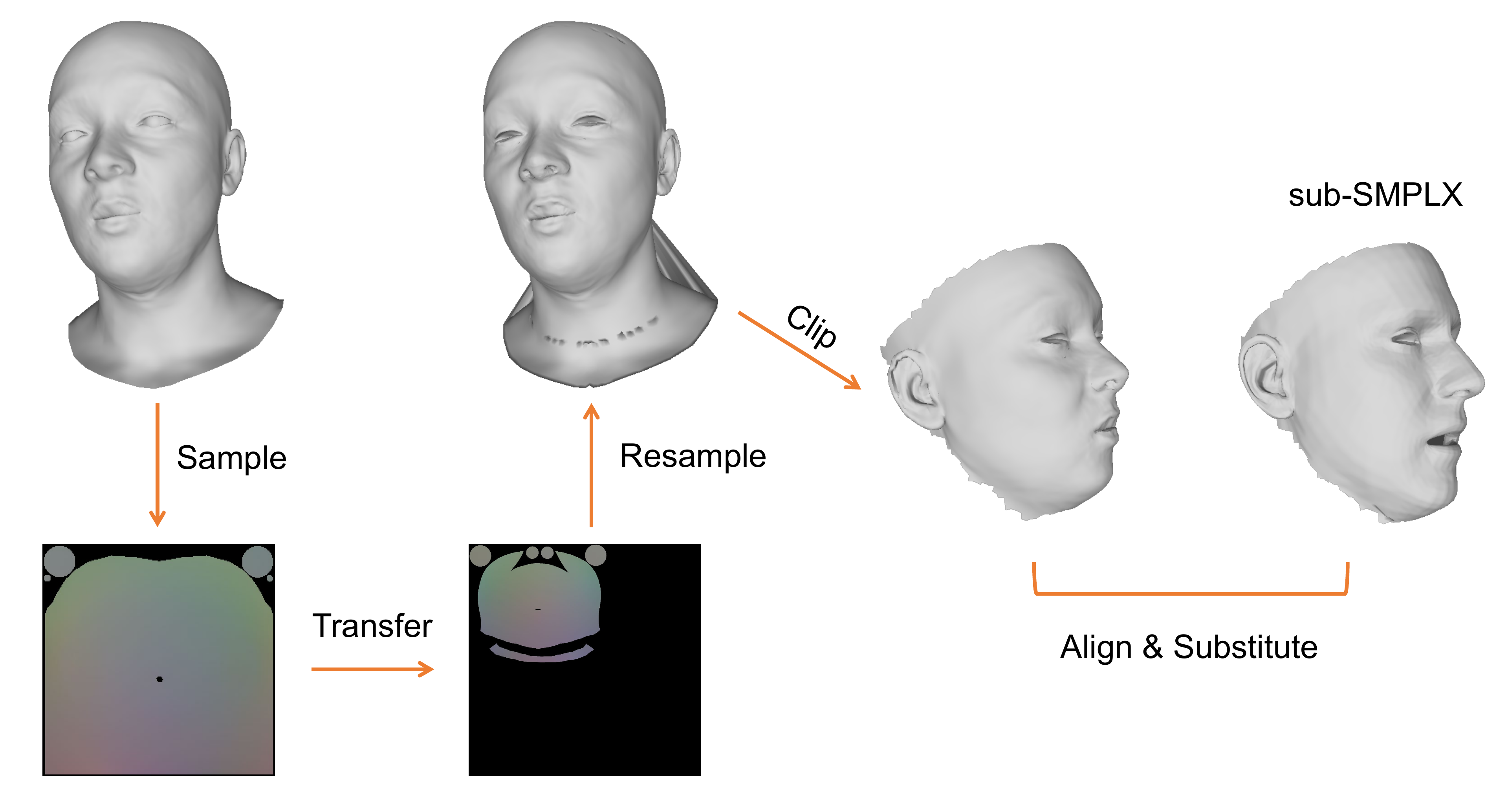}
   \vspace{-8mm}
   \caption{\textbf{Face Substitution}. We transform the detailed FLAME-based face model to sub-SMPLX representation to achieve face substitution.}
   \vspace{-5mm}
   \label{fig:face_substitution}
\end{figure}

\textbf{FLAME to SMPL-X}. We first transform the detailed faces of EMOCA \cite{EMOCA:CVPR:2021} to the UV domain since it has more vertices than the standard FLAME \cite{FLAME:2017:tog} model. We then transform the detailed FLAME UV position map to SMPL-X \cite{SMPL-X:2019} representation. The corresponding detailed sub-SMPLX face can be resampled from the transfered UV map. We clip the resampled face and align it with the face of sub-SMPLX, then performe vertex substitution. The whole process is shown in Fig. \ref{fig:face_substitution}.

\noindent
\textbf{Face Alignment}. We first rotate both the detailed face and sub-SMPLX face to align them with the positive z-axis, based on their respective average normal vectors in the x, y, and z directions. We then normalize them to a 0-1 space through translation and scaling. To achieve accurate alignment, we use the ICP algorithm \cite{icp}. Finally, we transform the aligned detailed face back into the coordinate space of the original sub-SMPLX. 

\begin{figure*}
  \centering
   \includegraphics[width=1\linewidth]{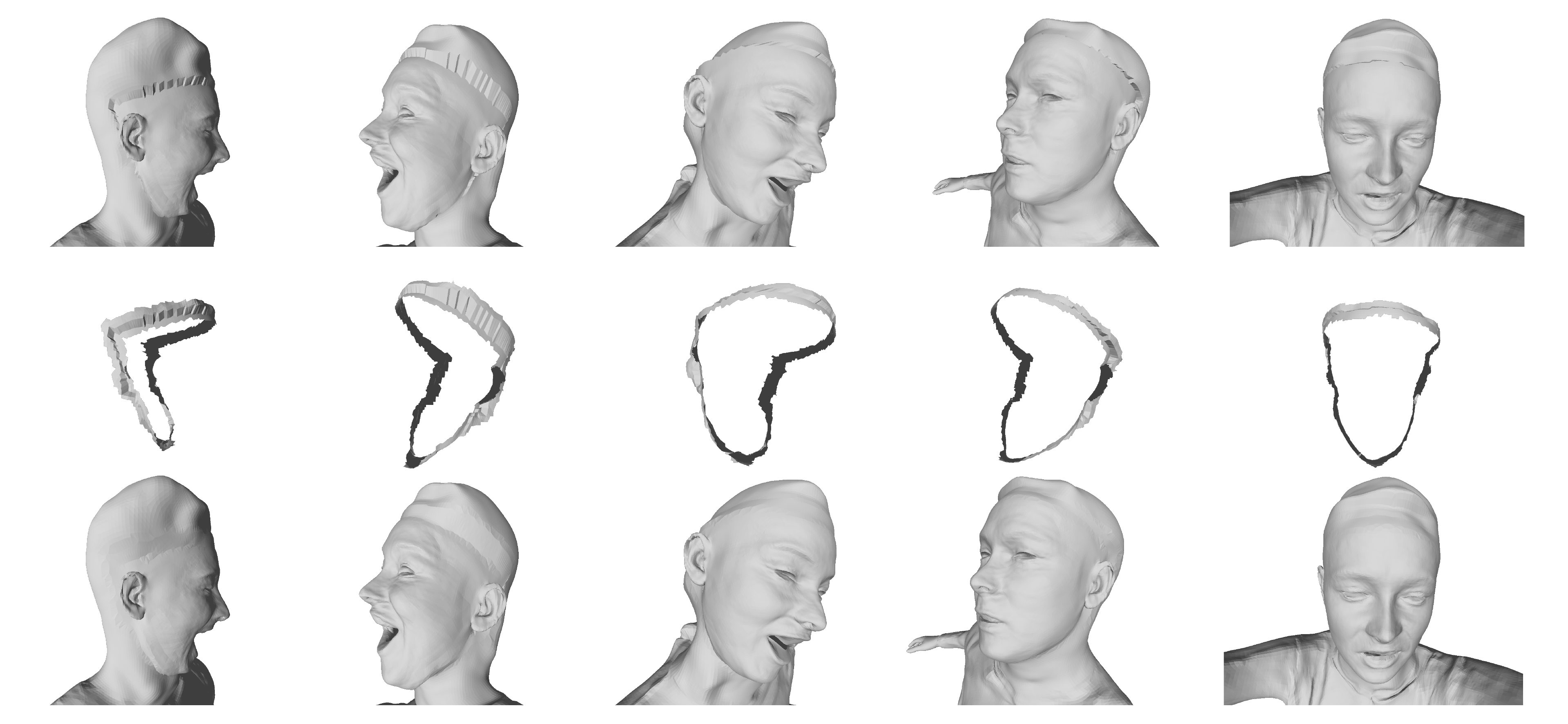}
   \caption{\textbf{Face edge smoothing}. The first line shows the results after face substitution, the second line shows the replaced edges, and the third line shows the smoothed results.}
   \label{fig:face_smooth}
\end{figure*}

\noindent
\textbf{Edge Smoothing}. Despite already aligning the detailed face and sub-SMPLX face through affine transformations, there is still a noticeable gap at the joint of the face and head. We resample the mesh between the face and the head using the corresponding UV position map and UV mask and reduce the gap at the joint through smoothing processing. The results are shown in Fig. \ref{fig:face_smooth}.

\subsection{Texture Diffusion}
\textbf{Partial Texture}.
We obtain the partial texture by projection (in Sec. \ref{a32}). In order to prevent the front colors from projecting onto invisible areas, we send a ray from each vertex along the camera's angle of view to detect the visible set. If the ray intersects with the body mesh, the corresponding vertex is considered invisible. Additionally, we use EMOCA \cite{EMOCA:CVPR:2021} to predict facial textures for better results.

\noindent
\textbf{Text Prompts}.
We use BLIP2 \cite{li2023blipv2} to generate text prompts corresponding to textures sampled from rendered images in Thuman2.0 for diffusion training.

\begin{figure*}
  \centering
   \includegraphics[width=1\linewidth]{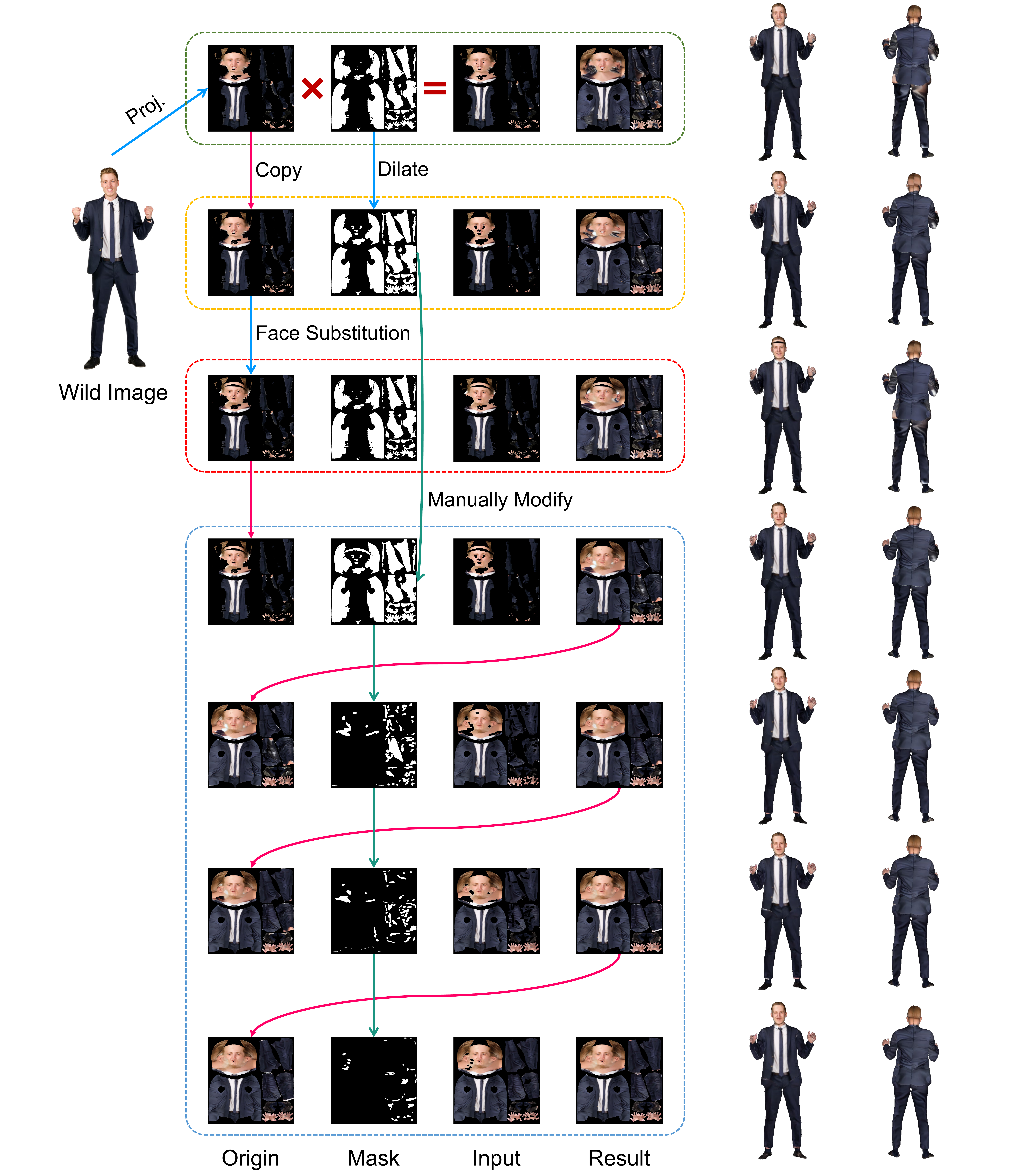}
   \caption{\textbf{Texture Inpainting}. We present four different strategies for inpainting the partial texture using SHERT. \textcolor{green}{Green}: Using the projected partial texture and mask directly. \textcolor{yellow}{Yellow}: Using the dilated mask. \textcolor{red}{Red}: Using the substituted facial texture. \textcolor{blue}{Blue}: Manual iterative optimization. The corresponding rendering results are displayed on the right.}
   \label{fig:texture_step}
\end{figure*}

\noindent
\textbf{Inference Details}.
We found inaccurate results at the edges of the projection, so we conduct dilation on the projected mask. Moreover, we use the facial texture predicted by EMOCA \cite{EMOCA:CVPR:2021} since the projected one frequently does not match the mesh. Lastly, we provide an iterative optimization scheme that allows users to manually recognize the sections to be optimized. The effects of different inpainting strategies are shown in Fig. \ref{fig:texture_step}. Sometimes the front and back of clothing may look similar, so we can also project the front color to the back of the mesh and conduct iterative optimization manually. This approach can achieve better results at the texture seams.

\section{Additional Experiments}
\label{sec:more}

\begin{figure*}
  \centering
   \includegraphics[width=1\linewidth]{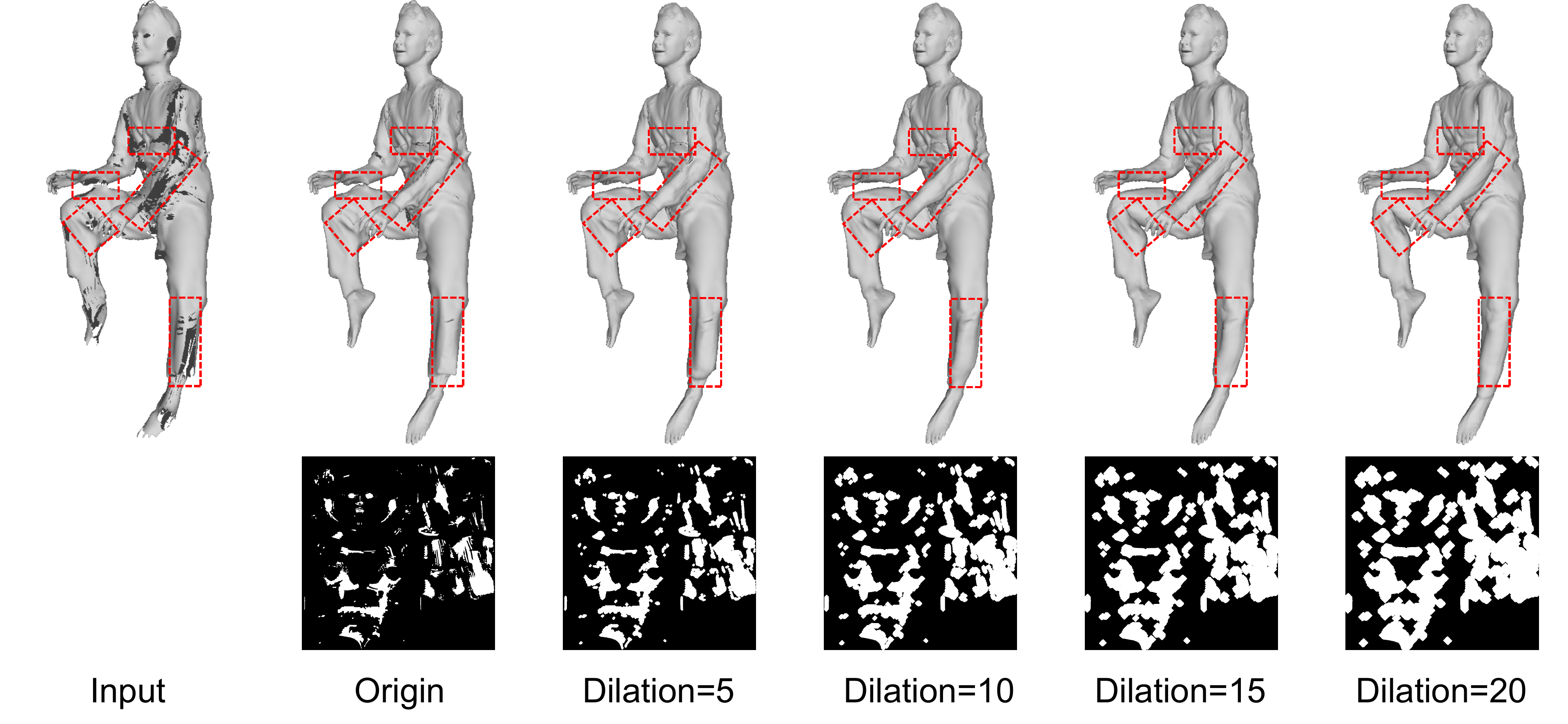}
   \caption{\textbf{Mask dilation in completion}. We display the completed results of different hole masks in the first line. The corresponding dilated hole masks are shown below.}
   \label{fig:dilate_completion}
\end{figure*}

\noindent
\textbf{The influence of mask dilation}.
We use mask dilation to reduce the impact of the edges of the culled meshes in Sec. \ref{a22}. As shown in Fig. \ref{fig:dilate_completion}, as the degree of dilation increases, the predicted results for the holed areas and its neighboring parts become smoother, but at the same time, more geometric details belonging to the original input are lost. 

\begin{figure*}
  \centering
   \includegraphics[width=1\linewidth]{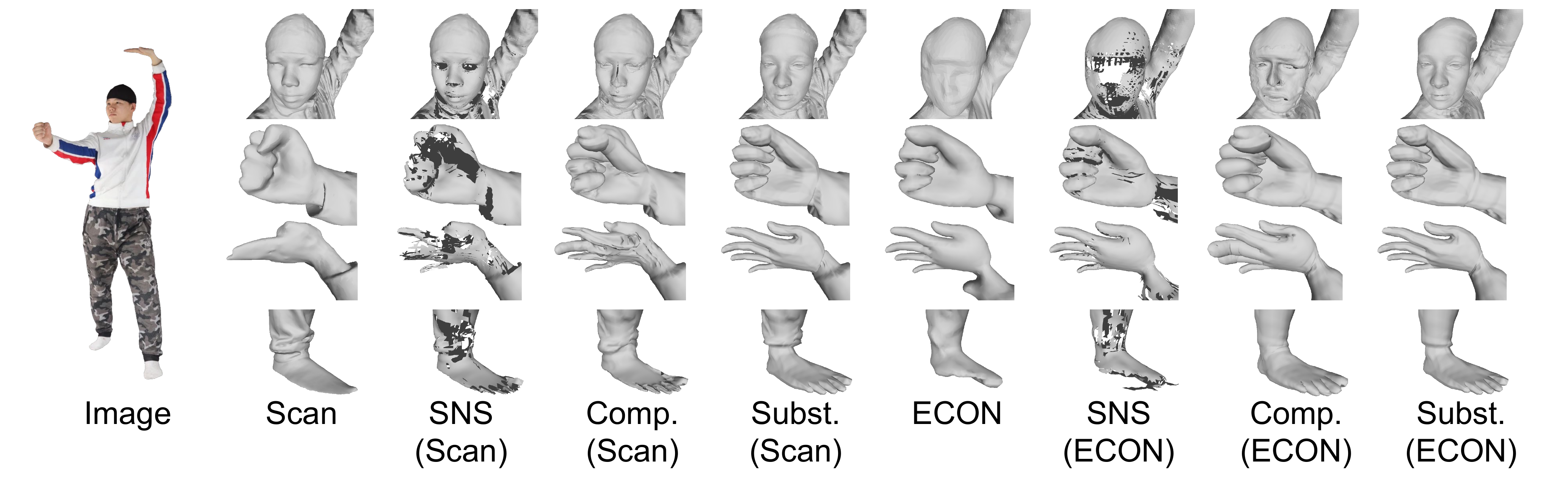}
   \caption{\textbf{The completion results of the face, hands, and feet}.}
   \label{fig:face_completion}
\end{figure*}

\noindent
\textbf{The completion results of the face, hands, and feet}.
The results are shown in Fig. \ref{fig:face_completion}

\begin{figure*}
  \centering
   \includegraphics[width=1\linewidth]{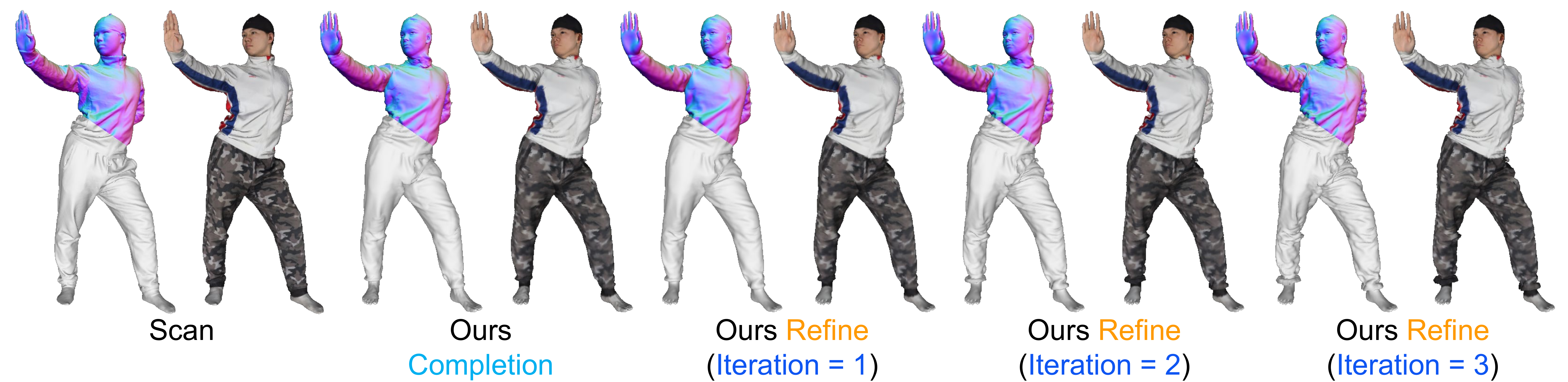}
   \caption{\textbf{The refine iteration for textured mesh}.}
   \label{fig:refine_texture}
\end{figure*}

\noindent
\textbf{The refine iteration for textured mesh}.
The results are shown in Fig. \ref{fig:refine_texture}

\begin{figure*}
  \centering
   \includegraphics[width=1\linewidth]{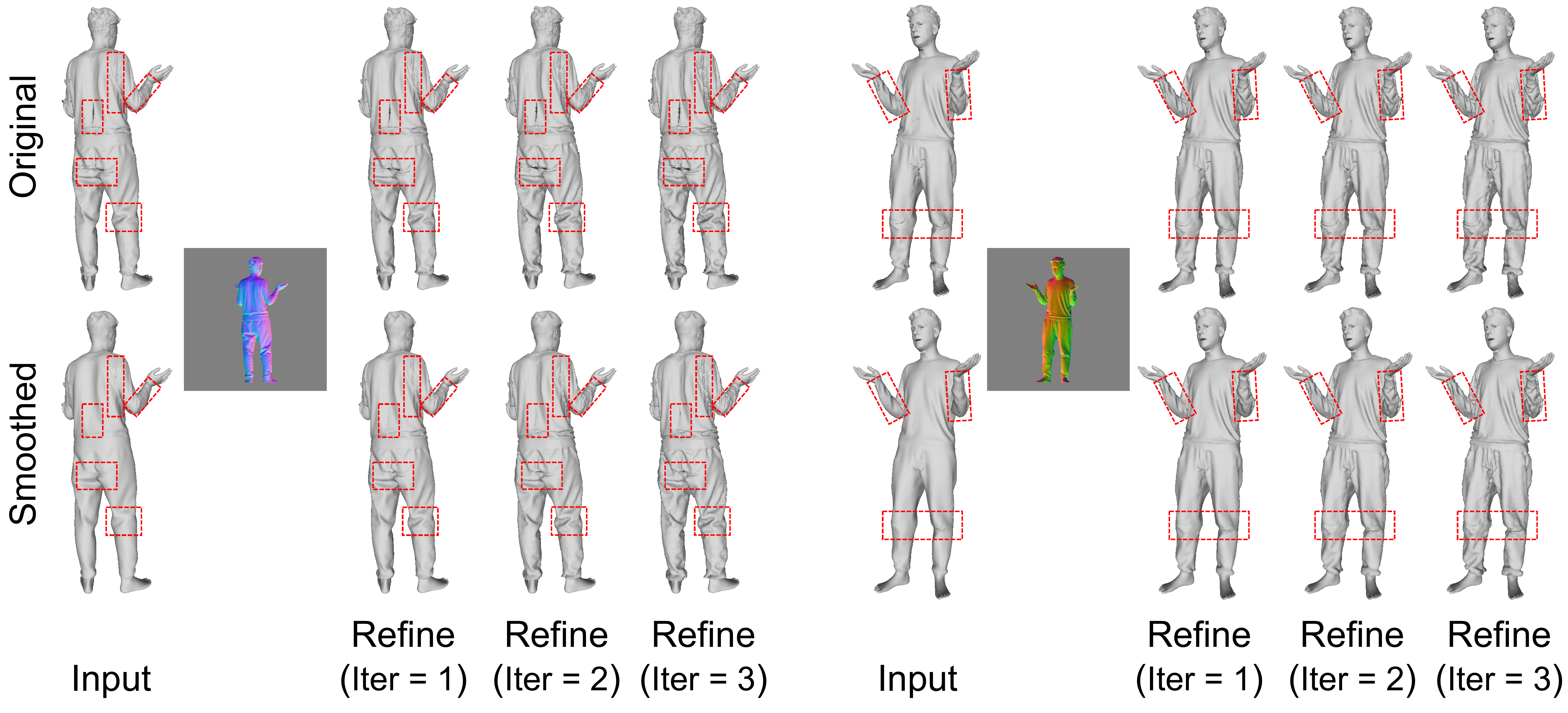}
   \caption{\textbf{The influence of smoothed inputs for refinement}. We show the refinement results of the original input in the first line and the smoothed input in the second line.}
   \label{fig:smooth_refine}
\end{figure*}

\noindent
\textbf{The influence of smoothed inputs for refinement}.
We demonstrate the effects of smoothed inputs for refinement in Fig. \ref{fig:smooth_refine}. The results indicate that while smoothing has minimal impact on the ultimate refinement of details, it is effective in curbing the generation of noise.

\begin{figure*}
  \centering
   \includegraphics[width=1\linewidth]{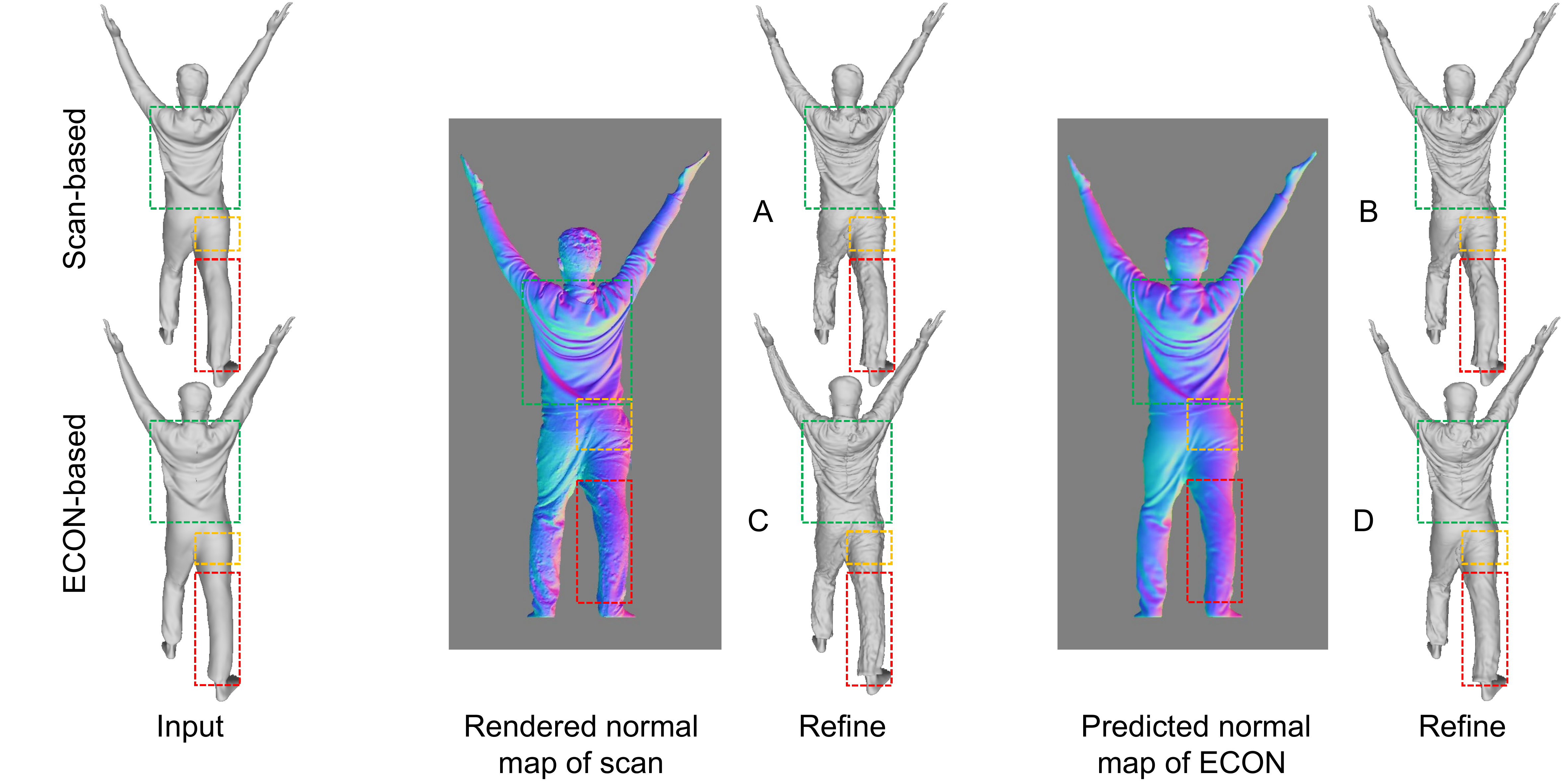}
   \vspace{-6mm}
   \caption{\textbf{The effects of normal quality}. We combine the mesh and normal obtained from the THuman2.0 scan, as well as the mesh and normal obtained through econ, to display four different results using the refinement network. The first line (A, B) uses the scan-based completed mesh as input, while the second line (C, D) uses the ECON-based completed mesh. The results shown on the left (A, C) utilize the scan-rendered normal map, while the ones on the right (B, D) utilize the ECON-predicted normal map. \textcolor{green}{Green}: When using matched normal maps (A, D), the enhancement of details is consistent with the geometric features of the model, resulting in stacked optimization results. However, when inconsistent inputs are used (B, C), the spatial position of the normal features does not correspond exactly to the geometric features of the mesh itself, resulting in a misalignment between the optimized details and the original geometric details. \textcolor{yellow}{Yellow}: When clear and detailed normal map (A, C) is inputted, the geometric details corresponding to the normal map can be optimized well, regardless of whether the input mesh matches the normal map. Conversely, if the details on the normal map are not clear enough (B, D), the details on the final optimized results will also be relatively blurred. \textcolor{red}{Red}: When there are noise in the normal map (A, C), the refinement network mistakenly adds this noise as details to the inputs. Conversely, a clean normal input (B, D) can effectively avoid such situation.}
   \label{fig:normal_quality}
\end{figure*}

\noindent
\textbf{The effect of normal quality in refinement}.
We train the refinement network with the normal maps of THuman2.0 \cite{thuman2:2021:cvpr} scans, which has a resolution of $1024 \times 1024$. However, the predicted normal maps of ECON \cite{econ2023} are $512 \times 512$. Such difference can lead to the degradation of refinement performance. Additionally, the quality of the normal and the degree of matching with the input geometry will also affect the final results. Experiments are carried out to show such effects on refine results, as shown in Fig. \ref{fig:normal_quality}.

\begin{figure*}
  \centering
   \includegraphics[width=1\linewidth]{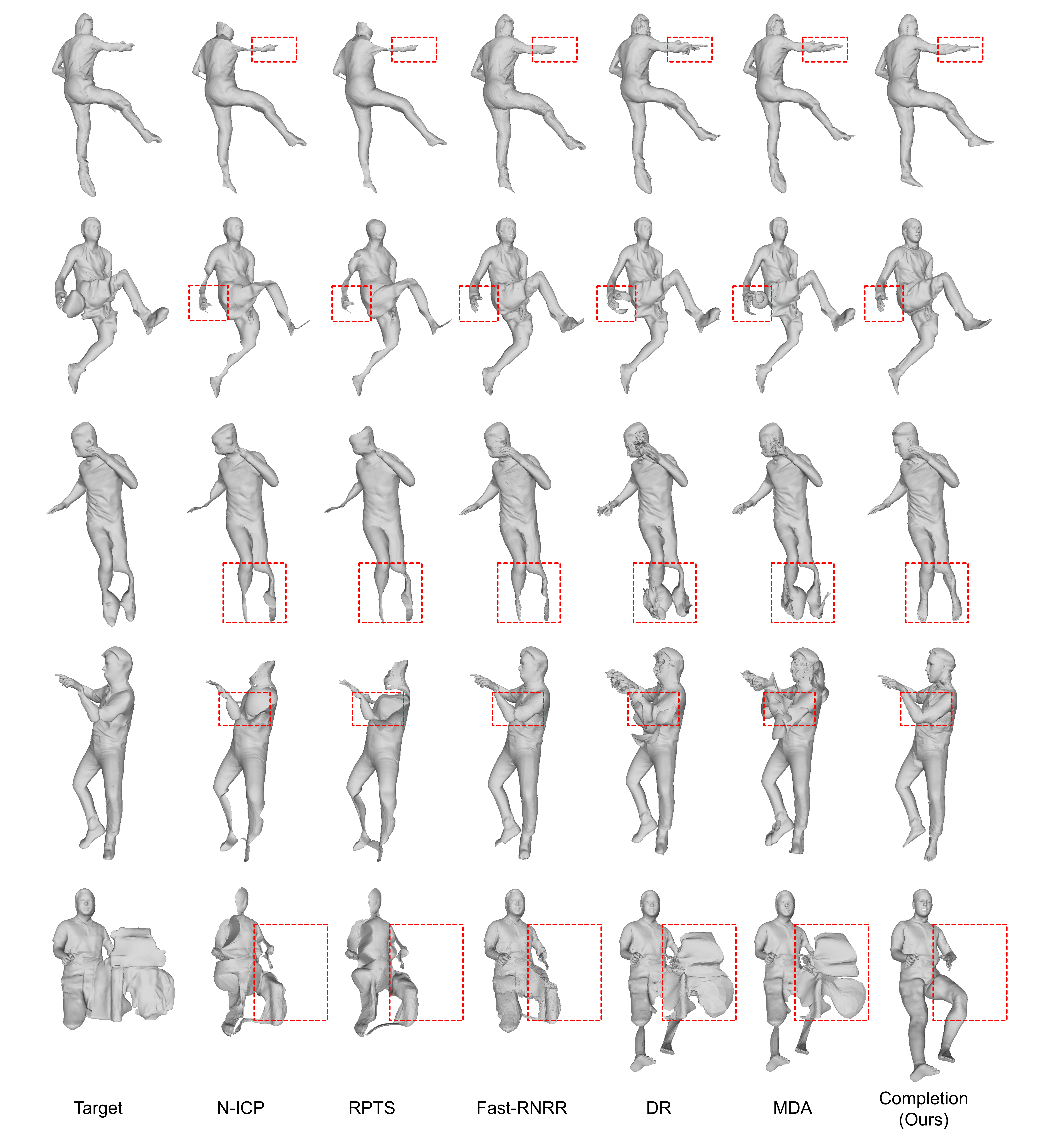}
   \caption{\textbf{Registration results for incomplete and incorrect inputs}. The results indicate that the SNS and completion processes of SHERT have significantly better robustness than other registration methods (N-ICP \cite{Nicp:2007}, RPTS \cite{RPTS}, Fast-RNRR \cite{fastRNRR:cvpr:2020}, DR \cite{Large:2021:sig}, MDA \cite{mda:2023:sig}).}
   \label{fig:incomplete_reg}
\end{figure*}

\noindent
\textbf{Registration on incomplete and incorrect inputs}.
The results are shown in Fig. \ref{fig:incomplete_reg}

\begin{figure*}
  \centering
   \includegraphics[width=1\linewidth]{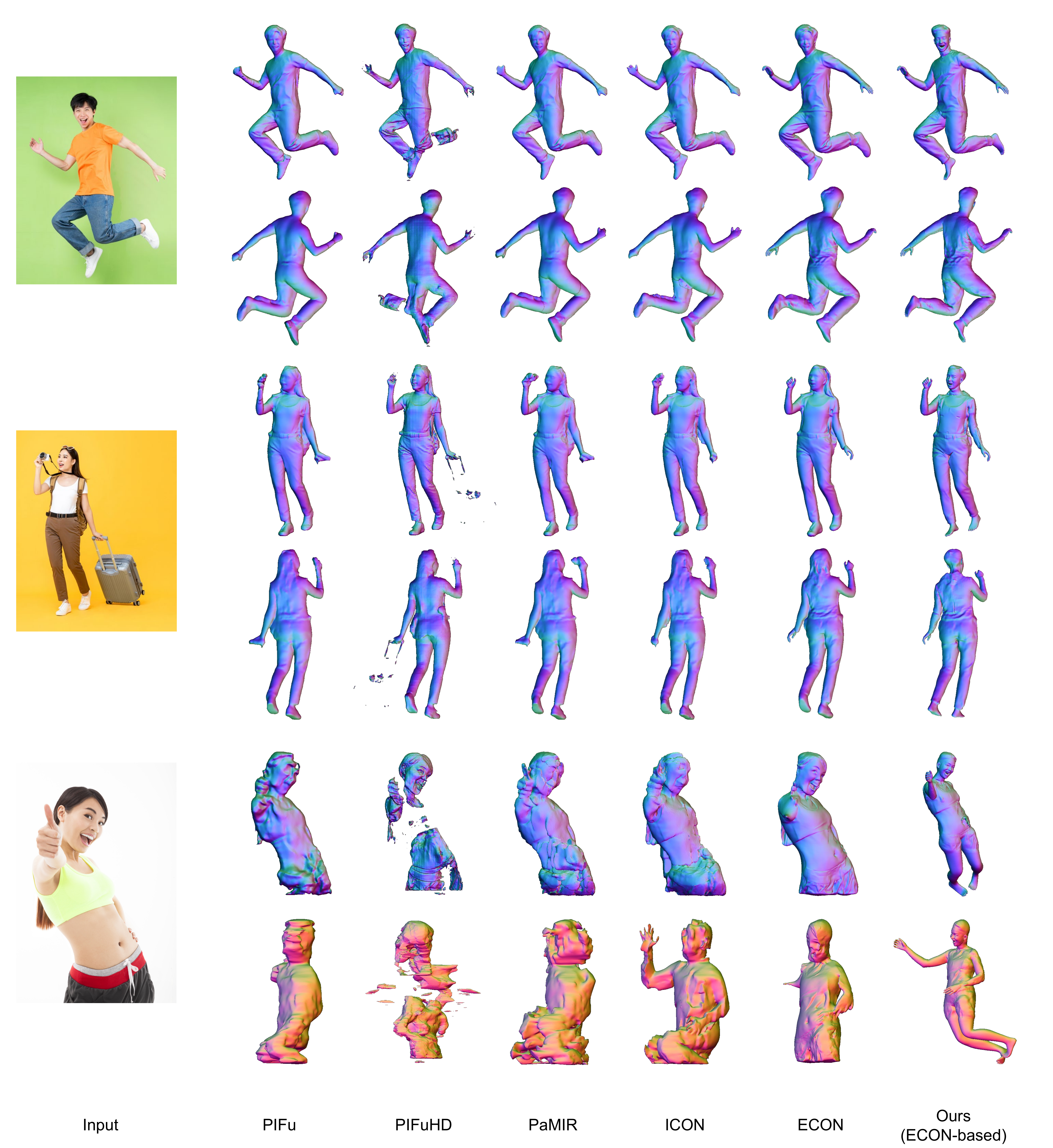}
   \caption{\textbf{Comparisons for monocular image reconstruction on in-the-wild images}. We compare the reconstruction quality of various methods including PIFu \cite{pifu:iccv:2019}, PIFuHD \cite{pifuhd:cvpr:2020}, PaMIR \cite{pamir:2021:pami}, ICON \cite{icon:2022:cvpr}, ECON \cite{econ2023}, and our SHERT. SHERT takes the ECON's result as the target surface. We mainly showcase the performance of SHERT when facing the complex pose (line 1-2), the long-haired person (line 3-4), and the incomplete input (line 5-6).}
   \label{fig:more_comparisons}
\end{figure*}

\noindent
\textbf{More comparisons}.
We present more comparisons for monocular image reconstruction in Fig. \ref{fig:more_comparisons}. We use challenging inputs to comprehensively evaluate the performance of SHERT.

\noindent
\textbf{More inpainting results}.
We present more inpainting results on in-the-wild images in Fig. \ref{fig:more_results}. The challenging inputs are also being used to demonstrate the performance unbiasedly.

\section{Limitations}
\label{sec:limi}

\noindent
\textbf{Incorrect sampling in SNS}.
SNS cannot guarantee that all of the unremoved faces are correct. When the target mesh is geometrically inseparable, SNS may incorrectly register a part of the body onto an incorrect surface.

\noindent
\textbf{Folded surfaces}.
Due to not using LBS \cite{lbs:2000:tog} for deformation during completion, but instead using normal-based offsets, there could be some folded surfaces, leading to discrepancies with the desired outcome.

\noindent
\textbf{Loose clothing, hair, and feet}.
Due to the limited expressiveness of the SMPL-X template we used, it is difficult for SNS to fully capture the geometry of loose clothing. Additionally, SHERT cannot accurately model long hair. Lastly, replacing the feet with the SMPL-X template resulted in an inability to reconstruct normal shoes in the final result.

\noindent
\textbf{Smoothed details}.
Some details of sampling failure during SNS may be lost during completion. At the same time, the smoothing used in refinement also smoothes out a small amount of detail.

\noindent
\textbf{Discontinuity of texture seams}.
Due to texture prediction being carried out on unfolded UVs, it is difficult to ensure consistency of the content at the edges of the seams, as they are not spatially adjacent.

\begin{figure*}
  \centering
   \includegraphics[width=1\linewidth]{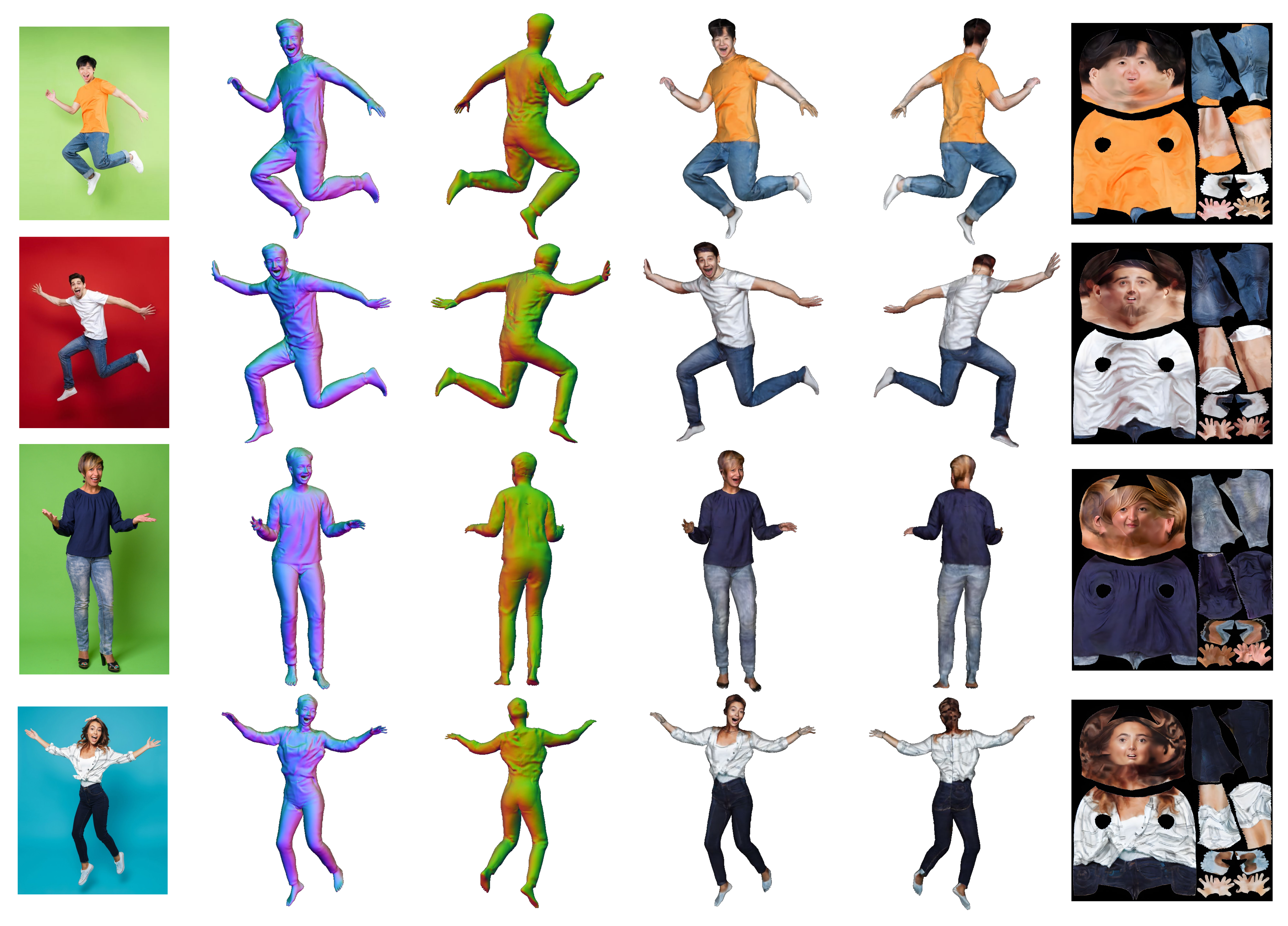}
   \caption{\textbf{Texture inpainting results on in-the-wild images}. }
   \label{fig:more_results}
\end{figure*}

\begin{figure*}
  \centering
   \includegraphics[width=1\linewidth]{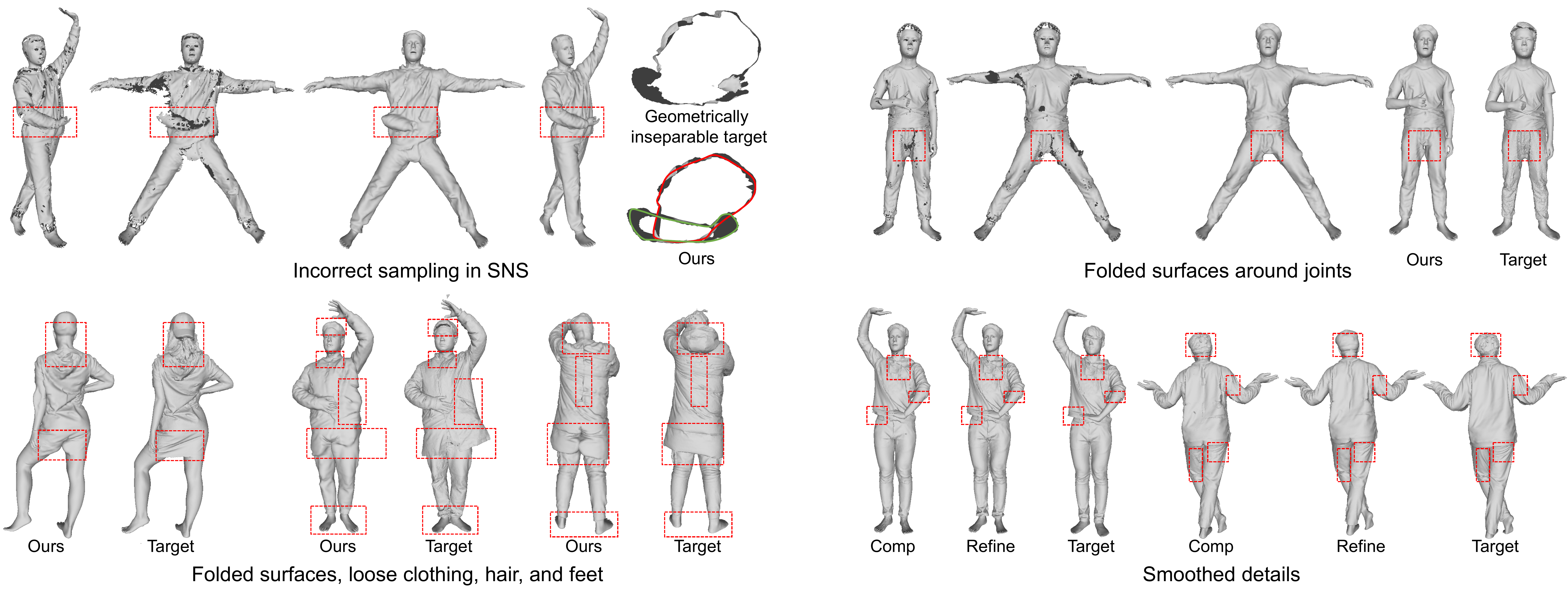}
   \caption{\textbf{Limitations}. }
   \label{fig:limitations}
\end{figure*}

The corresponding results are shown in Fig. \ref{fig:limitations} and Fig. \ref{fig:more_results}.

\section{Applications}
\label{sec:appl}

\begin{figure*}
  \centering
   \includegraphics[width=1\linewidth]{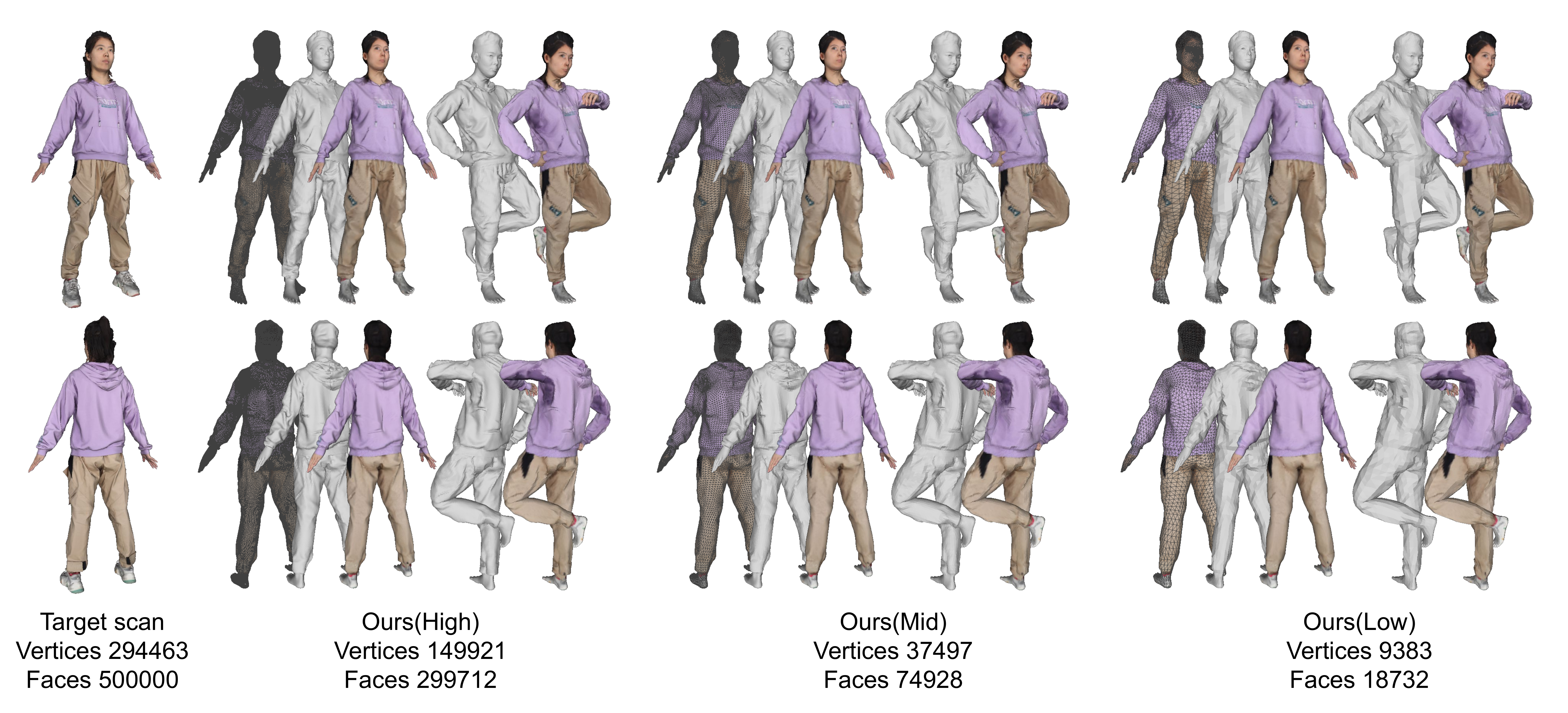}
   \caption{\textbf{Mesh down-sampling}. SHERT makes it possible to efficiently generate high-fidelity low-resolution models from a high-resolution model by using pre-defined human model templates.}
   \label{fig:downsample}
\end{figure*}

\noindent
\textbf{Mesh down-sampling}.
Because we have retained the original vertex ordering during the subdivision process, our representation supports direct down-sampling of the predicted results without any additional computation by using the pre-defined SMPL-X model (refer to Fig. \ref{fig:downsample}). The low-resolution results can preserve the geometric details, material and mesh quality well.

\begin{figure*}
  \centering
   \includegraphics[width=1\linewidth]{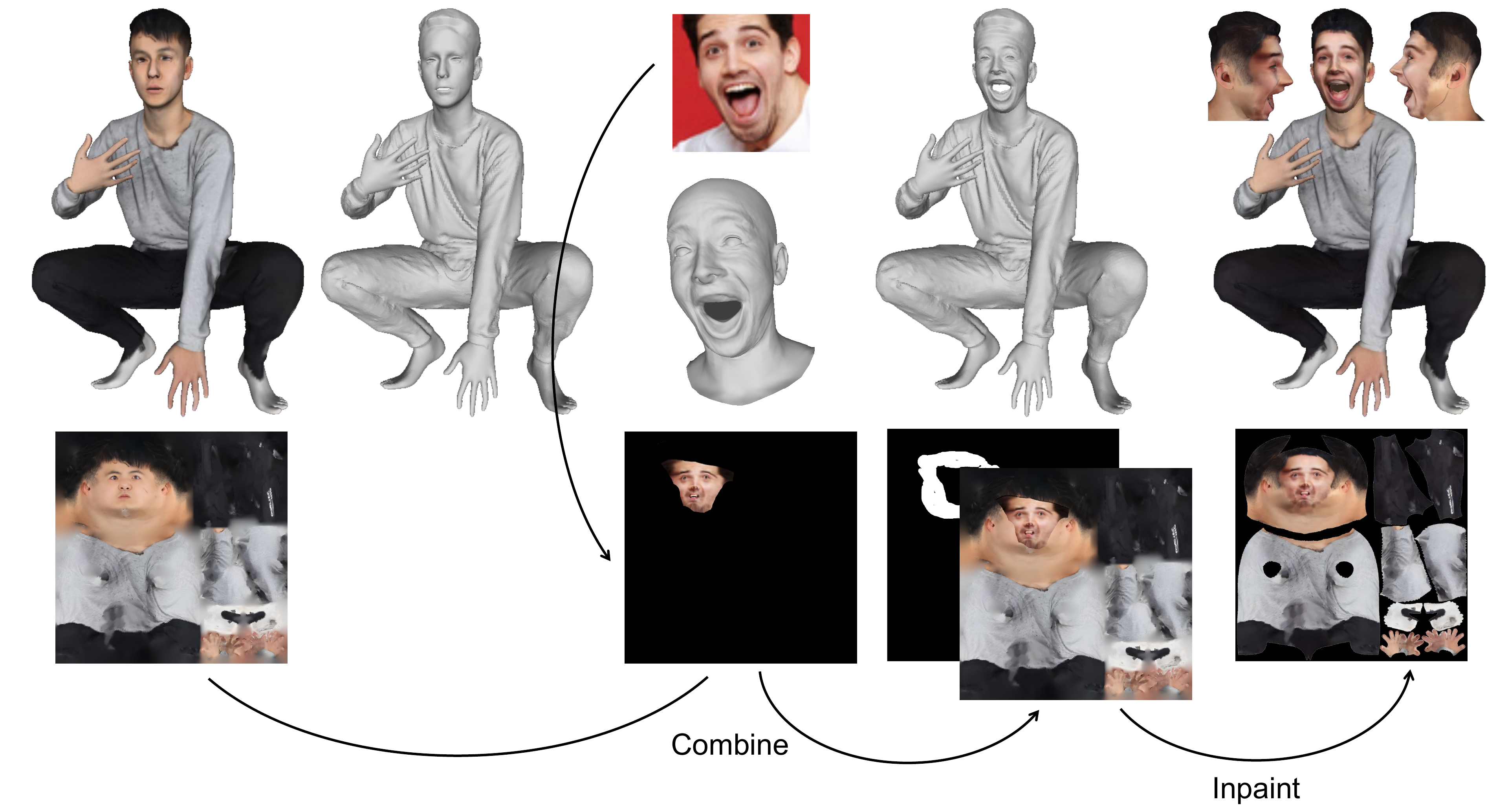}
   \caption{\textbf{3D face substitution}.}
   \label{fig:3d_face}
\end{figure*}

\noindent
\textbf{3D face substitution}. As shown in Fig. \ref{fig:3d_face}, SHERT enables direct 3D face substitution, allowing the face corresponding to the input image to be transferred to any existing mesh. With manual optimization, the facial texture can also be accurately transferred.

\clearpage


\end{document}